\documentclass{article}
\usepackage{spconf,amsmath,graphicx}
\usepackage{subfigure}
\usepackage[utf8]{inputenc}
\usepackage{overpic}
\usepackage{xcolor}
\usepackage{makecell}
\usepackage{booktabs}
\usepackage{tikz}
\usepackage{amssymb}
\usepackage{verbatim}
\usepackage{enumitem}
\usepackage[hyphens]{url}
\usepackage{hyperref}
\usepackage[outline]{contour}
\usepackage{tcolorbox}
\usepackage{soul}

\newcommand\copyrighttext{%
  \footnotesize \textcopyright 2022 IEEE. Personal use of this material is permitted.
  Permission from IEEE must be obtained for all other uses, in any current or future
  media, including reprinting/republishing this material for advertising or promotional
  purposes, creating new collective works, for resale or redistribution to servers or
  lists, or reuse of any copyrighted component of this work in other works. }
    \newcommand\mycopyrightnotice{%
\begin{tikzpicture}[remember picture,overlay]
\node[anchor=south,yshift=10pt] at (current page.south) {\fbox{\parbox{\dimexpr\textwidth-\fboxsep-\fboxrule\relax}{\copyrighttext}}};
\end{tikzpicture}%
}

\title{A Wavelet-based Dual-stream Network for Underwater \\Image Enhancement}

\name{Ziyin Ma and Changjae Oh}
\address{School of Electronic Engineering and Computer Science, Queen Mary University of London, UK}

\begin{document}
\ninept
\maketitle

\begin{abstract}
We present a wavelet-based dual-stream network that addresses color cast and blurry details in underwater images. We handle these artifacts separately by decomposing an input image into multiple frequency bands using discrete wavelet transform, which generates the downsampled structure image and detail images. These sub-band images are used as input to our dual-stream network that incorporates two sub-networks: the multi-color space fusion network and the detail enhancement network. The multi-color space fusion network takes the decomposed structure image as input and estimates the color corrected output by employing the feature representations from diverse color spaces of the input. The detail enhancement network addresses the blurriness of the original underwater image by improving the image details from high-frequency sub-bands. We validate the proposed method on both real-world and synthetic underwater datasets and show the effectiveness of our model in color correction and blur removal with low computational complexity.
\end{abstract}
\mycopyrightnotice

\begin{keywords}
Underwater image enhancement, wavelet decomposition, multi-color space, dual-stream network
\end{keywords}

\section{Introduction}
Underwater images and videos suffer from degradation due to the effects of light absorption and scattering, which causes blur and color casts. Image filtering tasks, e.g. restoration and enhancement, can improve the visual quality of underwater images, which can be divided into physics-based and learning-based methods~\cite{li2020underwatersurvey}.

\textit{Physics-based} methods build a model based on the physical and optical properties of the images taken in water~\cite{song2018rapid,peng2017underwater,drews2013transmission}. These methods investigate the physical mechanism of the degradation caused by color cast or scattering and compensate them to improve the underwater images. However, a single physics-based model cannot address all the complex physical and optical factors underlying the underwater scenes. This limitation leads to poor generalization and causes the results with over- or under-enhancement.

\textit{Learning-based} methods employ the representation power of deep neural networks (DNNs)~\cite{wang2019uwgan,li2017watergan,li2020underwater,anwar2018deep}. These methods commonly suffer from a lack of training data as paired underwater and clean images are difficult to collect. This issue can be addressed by generative adversarial network (GAN) based style transfer that does not require the paired training data and can transfer the appearance of a clean image to an underwater image~\cite{fabbri2018enhancing}. The training data can be augmented by generating synthetic images that include various color types, turbidity, and the illumination of water~\cite{li2020cast}.
This training strategy can further consider the diversity of water types in DNNs by adding a classifier and learning domain agnostic features to restore images~\cite{uplavikar2019all}.
Multi-color space can be employed to analyze the input image in various color domains, which allows the DNN to produce diverse feature representations~\cite{Ucolor}. Existing learning-based methods commonly focus on elaborating the network or training strategy, but they give less attention to the underlying complex artifacts to address in underwater images.

\begin{figure}[t]
\centering
\renewcommand{\arraystretch}{0.5}
\begin{tabular}{c@{\hspace{2pt}}c@{\hspace{2pt}}c}
\includegraphics[width=0.27\linewidth, height=0.055\textheight]{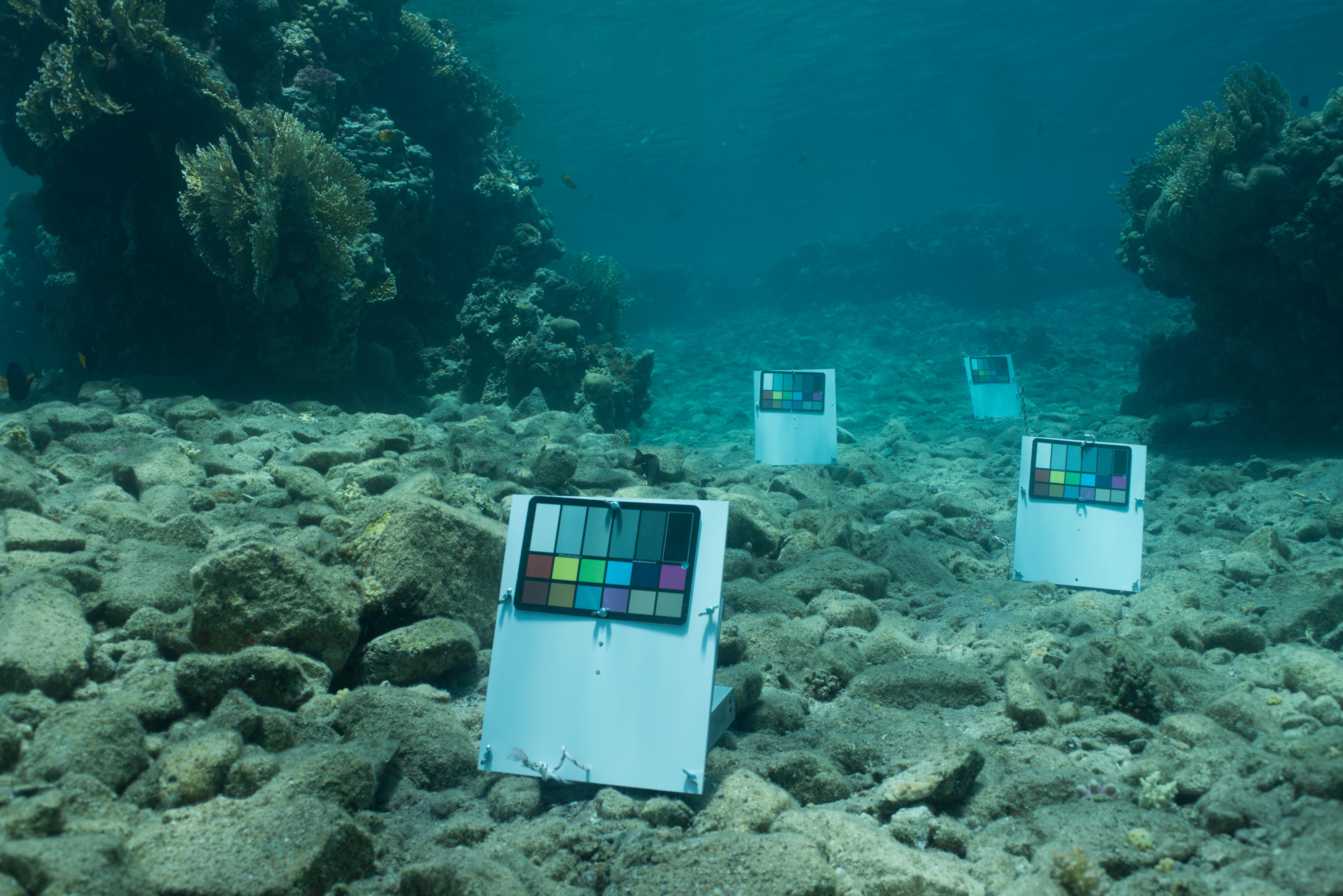}
&\includegraphics[width=0.27\linewidth, height=0.055\textheight]{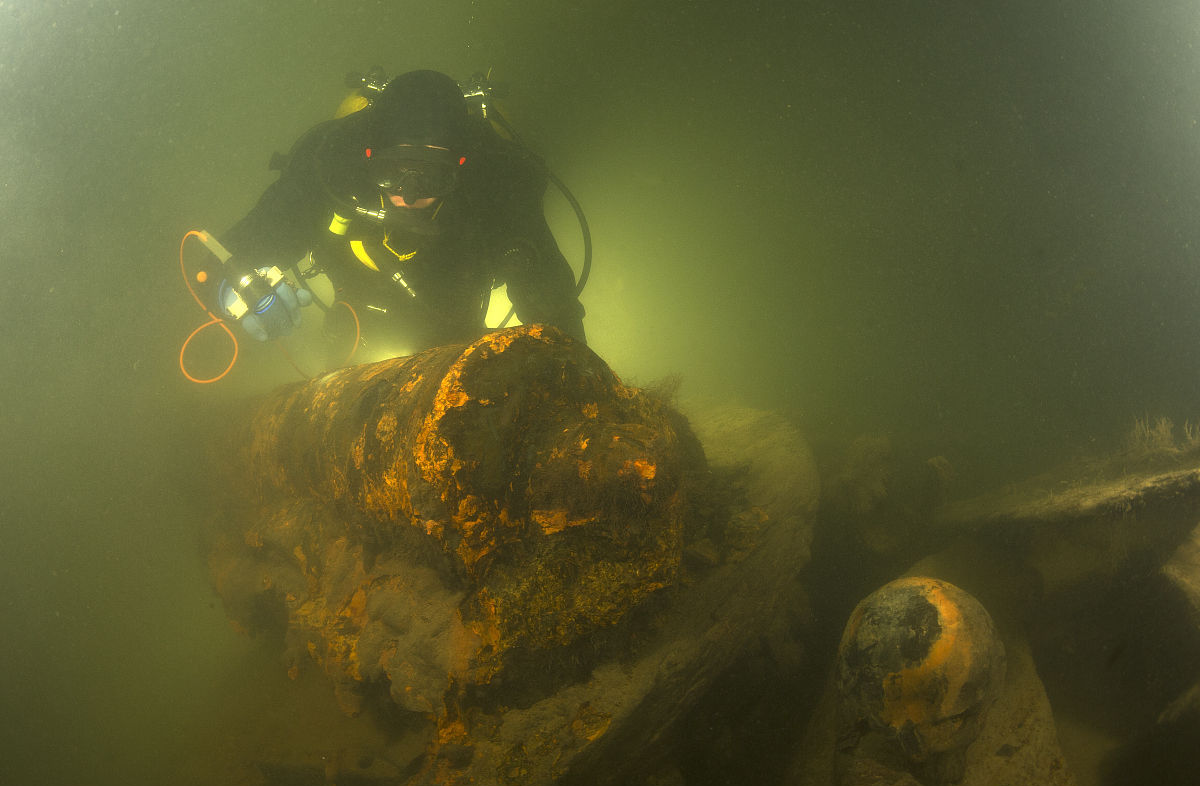}
&\includegraphics[width=0.27\linewidth, height=0.055\textheight]{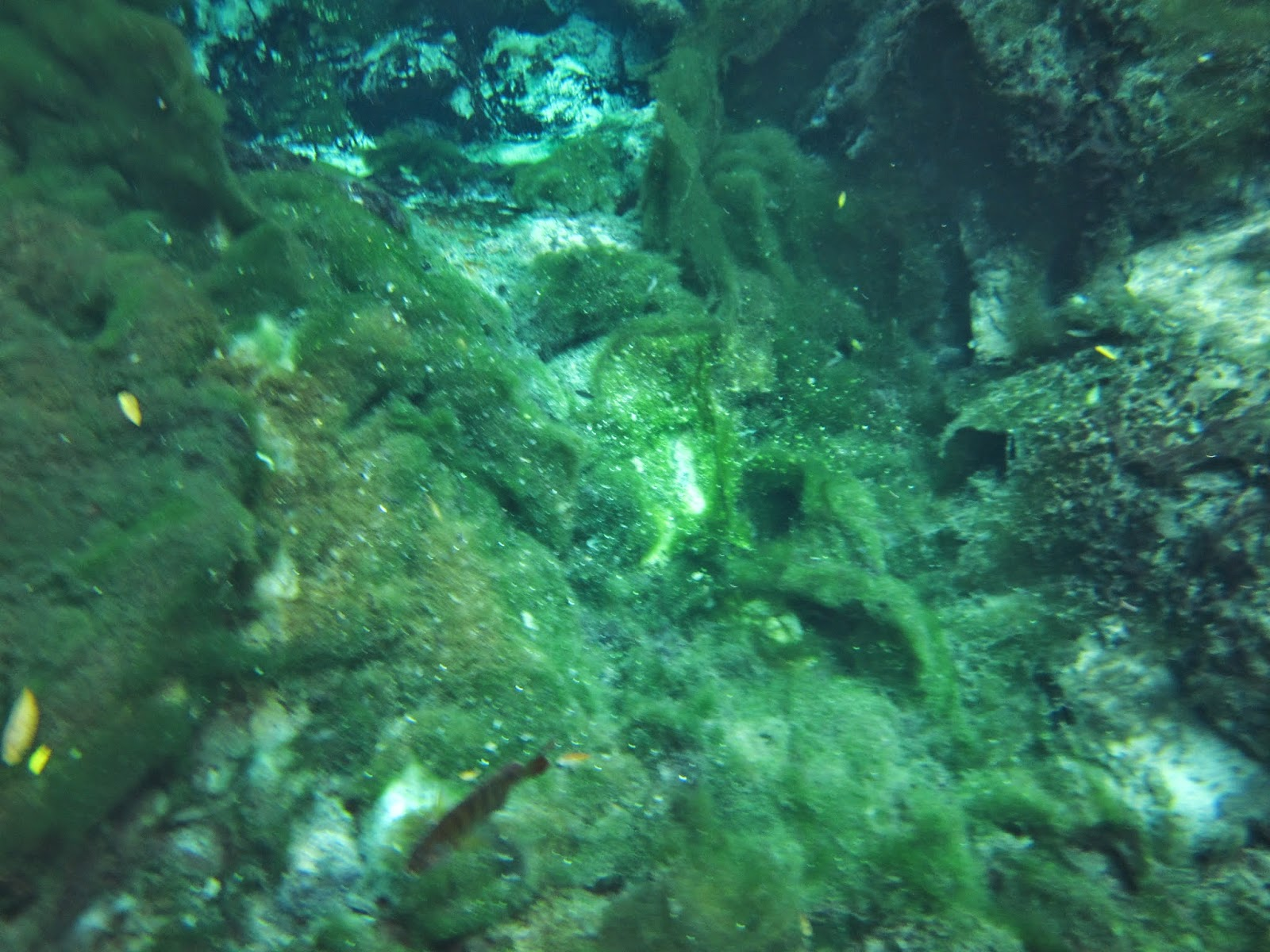}\\
\includegraphics[width=0.27\linewidth, height=0.055\textheight]{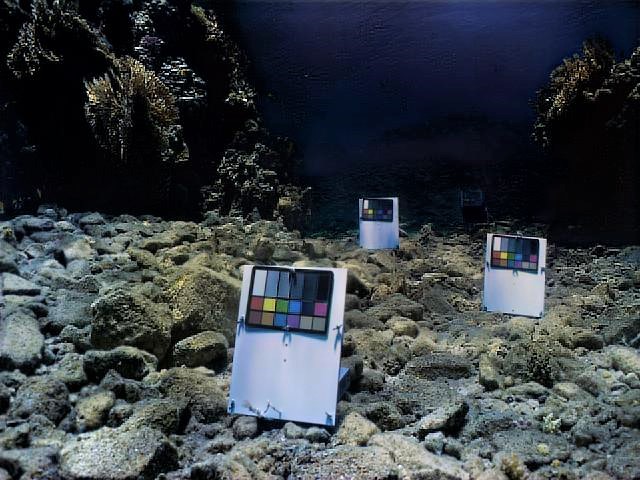}
&\includegraphics[width=0.27\linewidth, height=0.055\textheight]{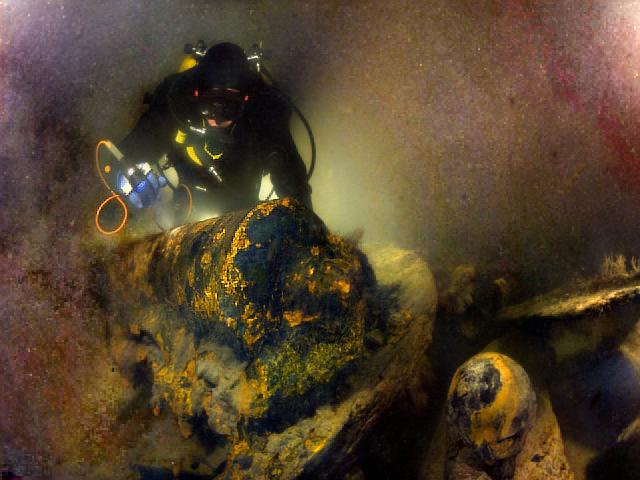}
&\includegraphics[width=0.27\linewidth, height=0.055\textheight]{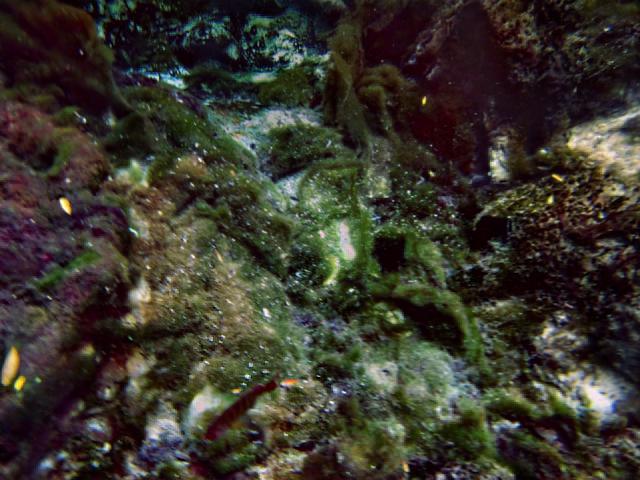}
\end{tabular}
\caption{(Top) Underwater images, which suffer from color cast and image blur, and (bottom) the results produced by ours.}
\vspace{-9pt}
\label{fig:example}
\end{figure}

In this paper, we propose a wavelet-based dual-stream network that performs the color correction and the detail enhancement separately by decoupling the problems of color casts and image blur in underwater images.
To this end, we first decompose an input image into sub-images with multiple frequency sub-bands using discrete wavelet transform, which enables each network to focus on addressing color casts and image blur, respectively. A sub-image with low-frequency contains the image structure at half-resolution while the rest of the sub-images contain image details. To deal with these sub-band images, we present a dual-stream network that addresses the color correction to the sub-image with low-frequency and the detail enhancement to the rest of the sub-images with higher frequency, respectively. Specifically, we present a multi-color space fusion network that considers the advantage of various color spaces of input for color-cast removal and a detail enhancement network to improve the missing details.
We validate our model on NYU-v2~\cite{silberman2012indoor}, UIEB~\cite{li2019underwater} and underwater color checker images (ColorChecker)~\cite{8058463}, which shows competitive results to existing methods. The main contributions of our work are as follows\footnote{{Results and code are available at: \url{https://zziyin.github.io/UIE-WD.html}} }:

\begin{itemize}[leftmargin=2em]
\setlength{\parskip}{1pt}
\item
We use wavelet decomposition to input underwater images, enabling our dual-stream network to perform the color correction and the detail enhancement, respectively.
\item We present a multi-color space fusion network that incorporates the various color representations for the color cast removal.
\item
Our model has the advantages in avoiding the artificial color and blurry outputs with low computational complexity.

\end{itemize}

\section{Proposed method}
\begin{figure}[t]
\includegraphics[width=\linewidth]{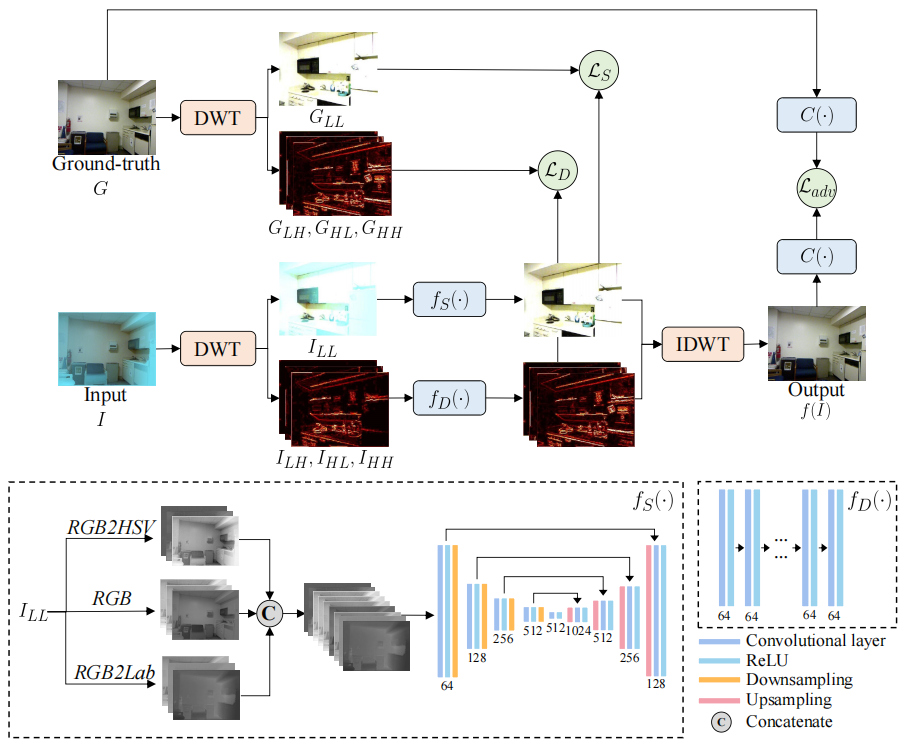}
\vspace{-9pt}
\vspace{-9pt}
\label{fig:network}
\caption{Overview of our method, trained on the synthetic dataset. The sub-band images with multiple frequency bands are obtained by discrete wavelet transform (DWT), which facilitates to decouple the color cast and blurry details in underwater images and separately address these artifacts with $f_{S}(\cdot)$ and $f_{D}(\cdot)$, respectively. $f_{S}(\cdot)$ and $f_{D}(\cdot)$ are constrained by the structure loss, $\mathcal{L}_{S}$, and the detail loss, $\mathcal{L}_{D}$, respectively. The output, $f(I)$, after inverse DWT (IDWT), is constrained by the adversarial loss, $\mathcal{L}_{adv}$, with the generative adversarial network (GAN) discriminator, $C(\cdot)$.}
\vspace{-9pt}
\end{figure}

Given an underwater image, $I$, we aim to learn a network, $f(\cdot)$, to generate an enhancement output, $f(I)$, that removes the color cast from $I$ while enhancing the image details. We first use discrete wavelet transform (DWT) to decompose $I$ into sub-band images that consist of an approximated original image at half-resolution, $I_{LL}$, and the images with high-frequency components into vertical $I_{LH}$, horizontal $I_{HL}$, and diagonal $I_{HH}$ directions. These sub-band images are then fed to the dual-stream network, i.e. the multi-color space fusion network, $f_S(\cdot)$, and the detail enhancement network, $f_D(\cdot)$, in which the color-cast removal and detail enhancement are separately processed. $f_S(\cdot)$ aims to remove the color cast in $I_{LL}$ by representing the input in the multiple color spaces, while $f_D(\cdot)$ aims to generate the detail enhancement output from $I_{LH}, I_{HL}$, and $I_{HH}$.
These sub-band estimations are then integrated and reconstructed to the original size using inverse DWT (IDWT). We jointly train our model with the structure loss, $\mathcal{L}_S$, detail loss, $\mathcal{L}_D$, and adversarial loss, $\mathcal{L}_{adv}$. Due to the lack of clean-underwater image pairs, the model is trained on the synthetic dataset~\cite{anwar2018deep,silberman2012indoor}. Fig.~\ref{fig:network} shows the overall pipeline of our method.

\subsection{Discrete wavelet decomposition}
Wavelet transform has been applied to various image processing tasks such as image super-resolution~\cite{huang2017wavelet} and denoising~\cite{kang2018deep}.
Several physics-based methods in underwater image enhancement have used DWT to decompose the images and process them in the frequency domain~\cite{singh2014enhancement, priyadharsini2018wavelet} to improve the contrast and resolution. In our framework, we use DWT to decompose an input image into multiple frequency sub-bands so that the color correction and detail enhancement can be separately addressed.

We decompose an input using Haar wavelets that consist of the low-pass filter, $L$, and the high-pass filter, $H$, as follows:
\begin{equation}
L=\frac{1}{\sqrt{2}}[1,1]^T, H=\frac{1}{\sqrt{2}}[1,-1]^T.
\end{equation}
We obtain four sub-band images $I_{LL}, I_{LH}, I_{HL}, I_{HH}$  by conducting convolution and downsampling on image $I$. $I_{LL}$ is obtained by using a low-pass filter $LL^T$ to horizontal and vertical directions. The other three sub-band images, $I_{LH}, I_{HL}, I_{HH}$, are obtained by using the filters $LH^T, HL^T, HH^T$, that captures the high-frequency components in vertical, horizontal, and diagonal directions.
In addition, the sub-band images are downsampled to half-resolution of the original input but do not result in information loss due to the biorthogonal property of DWT. DWT can be seen as a convolution process on $I$, using four $2 \times 2$ convolution kernels with fixed weights and with the stride of 2, and IDWT can be seen as the transposed convolution. The proposed network thus can be trained end-to-end.

\begin{figure}[t]
\centering
\renewcommand{\arraystretch}{0.5}
\begin{tabular}{c@{\hspace{2pt}}c@{\hspace{2pt}}c@{\hspace{2pt}}c@{\hspace{2pt}}c}
\includegraphics[width=0.18\linewidth]{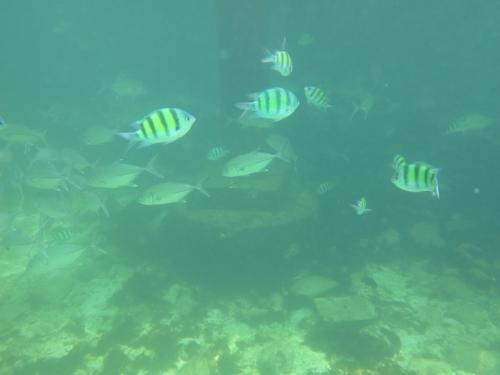}
&\includegraphics[width=0.18\linewidth]{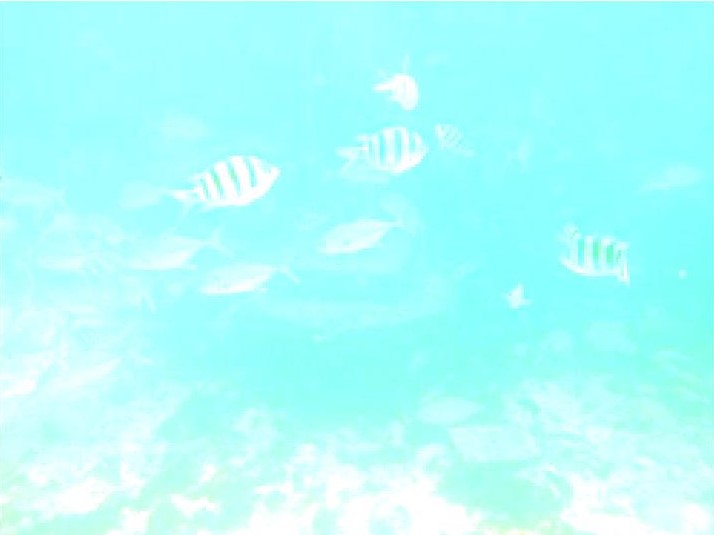}
&\includegraphics[width=0.18\linewidth]{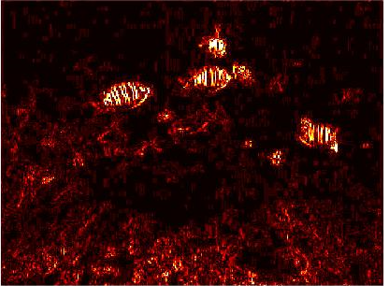}
&\includegraphics[width=0.18\linewidth]{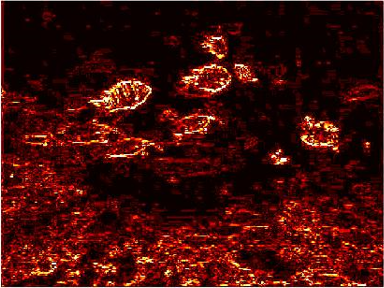}
&\includegraphics[width=0.18\linewidth]{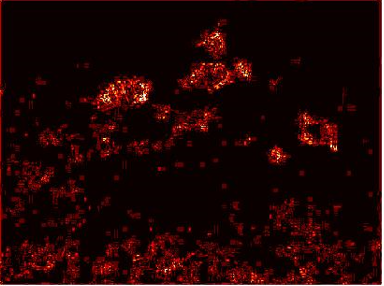}\\
\includegraphics[width=0.18\linewidth]{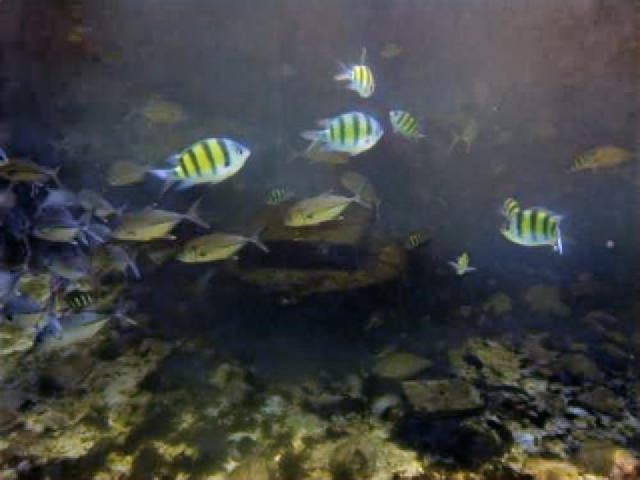}
&\includegraphics[width=0.18\linewidth]{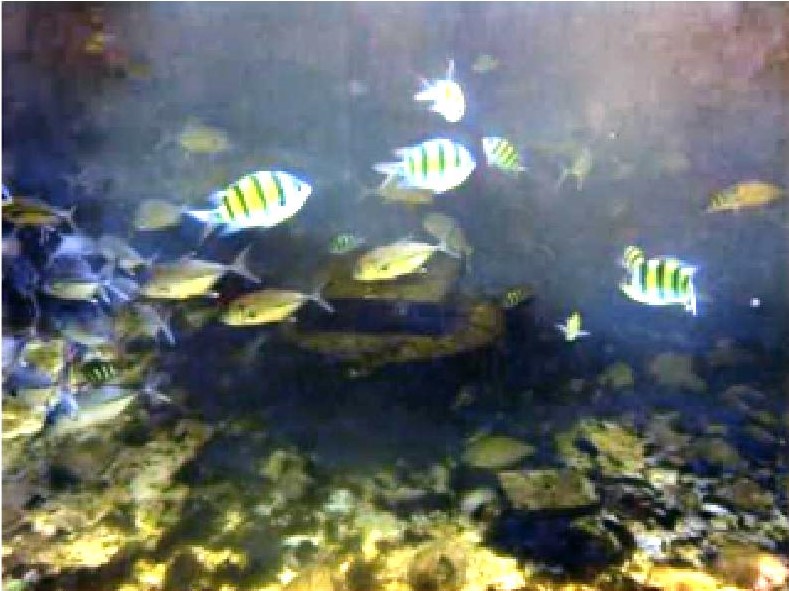}
&\includegraphics[width=0.18\linewidth]{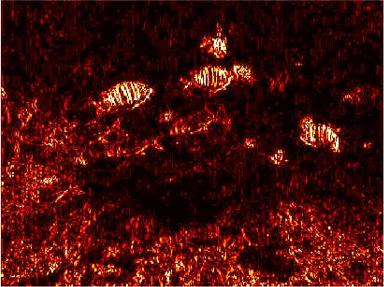}
&\includegraphics[width=0.18\linewidth]{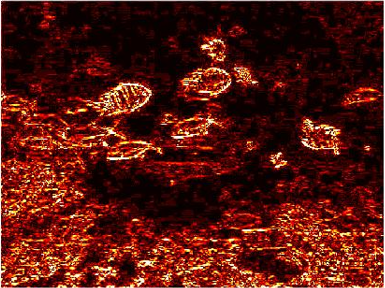}
&\includegraphics[width=0.18\linewidth]{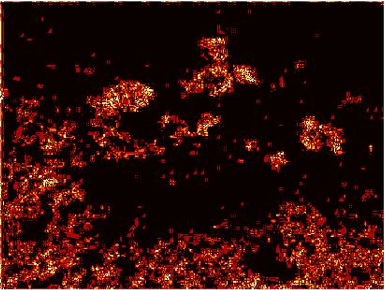}\\
\footnotesize{Input}
&\footnotesize{$I_{LL}$}
&\footnotesize{$I_{LH}$}
&\footnotesize{$I_{HL}$}
&\footnotesize{$I_{HH}$}
\end{tabular}
\caption{Decomposition results of the (top) underwater image and (bottom) enhancement output after 1-scale DWT. Note that input images are downsampled to the half resolution, and the normalized absolute values are color-coded for $I_{LH}$, $I_{HL}$ and $I_{HH}$, where the pixel value increases from black to red to yellow.}
\label{fig:decom}
\vspace{-9pt}
\end{figure}

Fig.~\ref{fig:decom} shows the original images and their sub-band images obtained by DWT.
We can observe that $I_{LL}$ includes most of the global image structure with the color cast, at the half-resolution, while $I_{LH}, I_{HL}, I_{HH}$ contain image details captured from different directions. To fully exploit these sub-band images by considering their property, we present a dual-stream network that separately processes the sub-band images for enhancement.

\subsection{Dual-stream network}
The proposed dual-stream network consists of two sub-networks, multi-color space fusion network, $f_S(\cdot)$, and detail enhancement network, $f_D(\cdot)$, which separately estimate the structure and detail component of wavelet sub-images, respectively.

\begin{figure}[t]
\centering
\renewcommand{\arraystretch}{0.5}
\begin{tabular}{c@{\hspace{2pt}}c@{\hspace{2pt}}c@{\hspace{2pt}}c}
\includegraphics[width=0.2\linewidth]{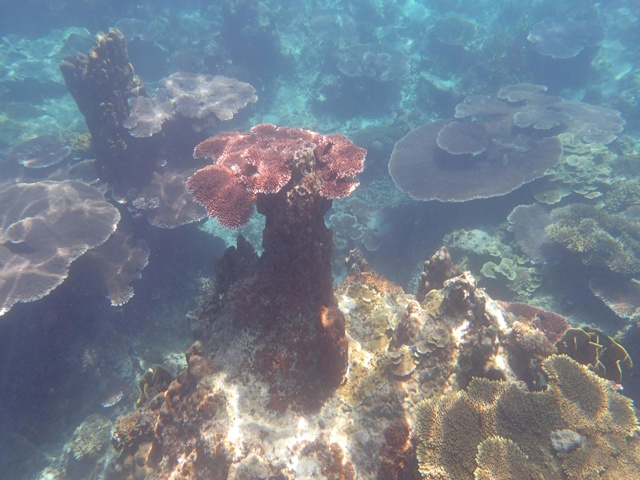}
&\includegraphics[width=0.2\linewidth]{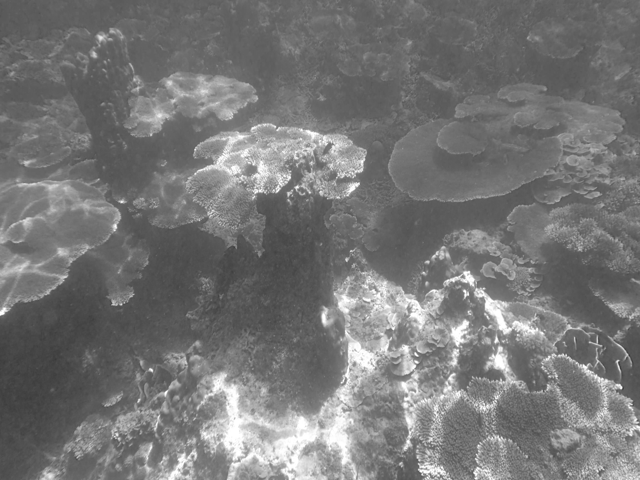}
&\includegraphics[width=0.2\linewidth]{figures/proposed method/HSVLAB/R.jpg}
&\includegraphics[width=0.2\linewidth]{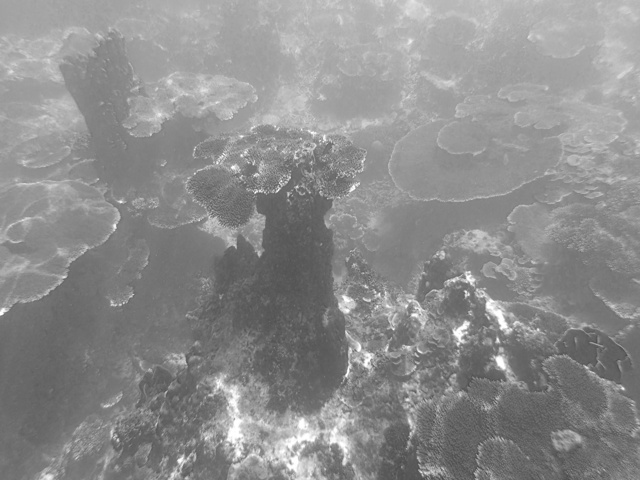}
\\

&\includegraphics[width=0.2\linewidth]{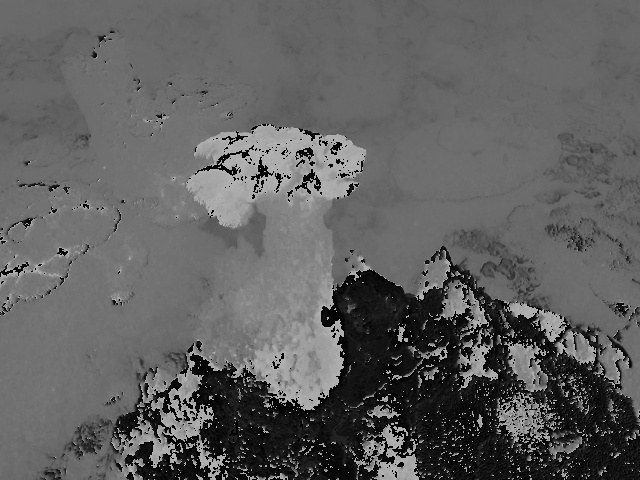}
&\includegraphics[width=0.2\linewidth]{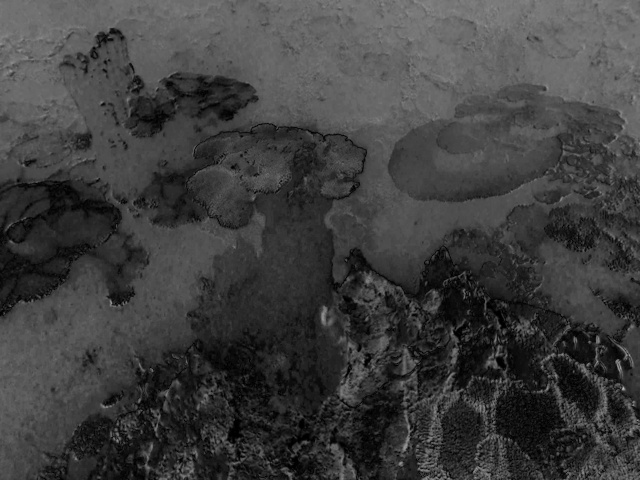}
&\includegraphics[width=0.2\linewidth]{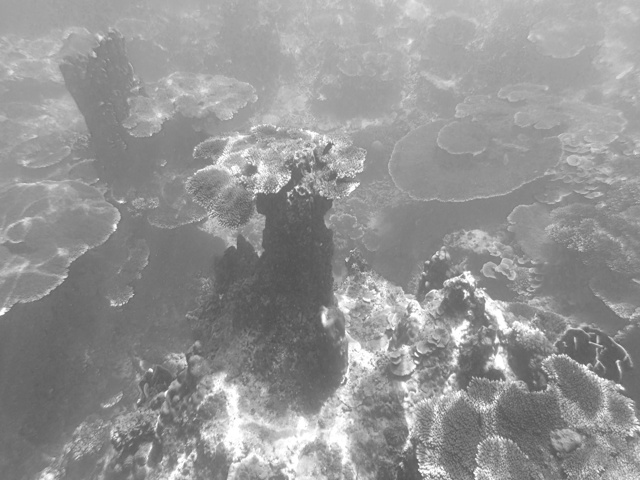}
\\
&\includegraphics[width=0.2\linewidth]{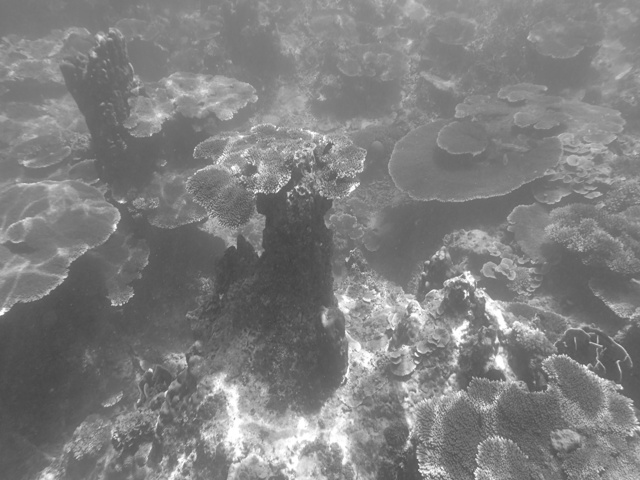}
&\includegraphics[width=0.2\linewidth]{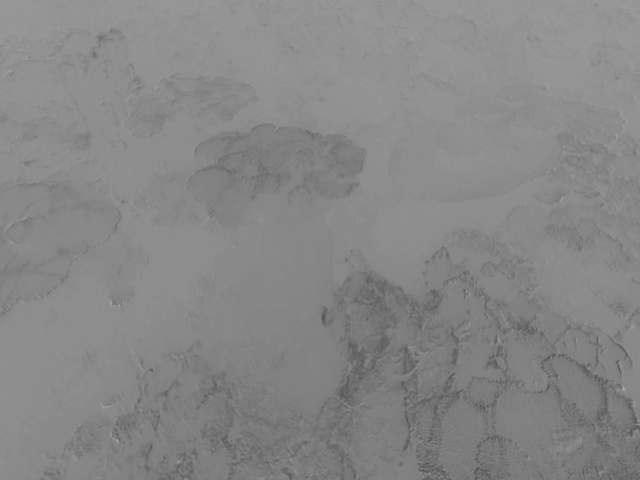}
&\includegraphics[width=0.2\linewidth]{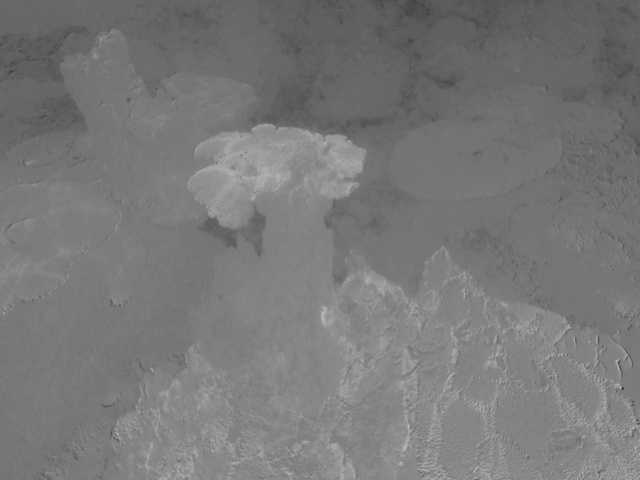}
\end{tabular}
\vspace{-9pt}
\caption{Images from different color spaces. (Top) An RGB image and its $R$, $G$, $B$, (middle) $H$, $S$, $V$, and (bottom) $L$, $a$, $b$ components. The darker the intensity, the smaller the value. }
\label{fig:HSV}
\vspace{-9pt}
\end{figure}

For $f_S(\cdot)$, we adopt U-net~\cite{ronneberger2015u} as our base architecture. Considering that the diverse optical factors in underwater cause the various color casts, we present a multi-color space fusion module to analyze the input image with various color representations.
In addition to using the common RGB color space for an input image, we transform the RGB input to HSV and Lab color spaces and then concatenate them into a 9-channel image to further extract diverse feature representations. As shown in Fig.~\ref{fig:HSV}, the HSV color space can directly reflect the brightness and contrast of the image. The Lab color space approximates the human visual system and provides perceptual uniformity~\cite{Ucolor}. We then pass the concatenated image with the multi-color spaces to the network to estimate the clean image without color cast.  Moreover, such non-linear color transforms can greatly improve the performance of the network without making the network deeper~\cite{Ucolor}

For $f_D(\cdot)$, we aim to address the image blur by decoupling the input from the color cast and focusing only on the image details to improve. The inputs with high-frequency bands, $I_{LH}$, $I_{HL}$, and $I_{HH}$, pass $f_D(\cdot)$ that consists of 10 convolution layers followed by the ReLU function. The filter size and the depth of each layer are $3\times3$ and 64, respectively. With $f_D(\cdot)$, the missing details can be improved as shown in Fig.~\ref{fig:decom}.

After reconstructing the final output $f(I)$ from $f_S(I_{LL})$, $f_D(I_{LH})$, $f_D(I_{HL})$ and $f_D(I_{HH})$ using IDWT, we further perform the adversarial learning for higher accuracy~\cite{arjovsky2017wasserstein}.
We add a GAN discriminator, $C(\cdot)$, which is trained to distinguish the ground truth $G$ from $f(I)$. This GAN framework encourages our network to generate more realistic images to mislead $C(\cdot)$.

\subsection{Loss functions}
We train the network with the loss function, $\mathcal{L}_{total}$, that combines the structural loss, $\mathcal{L}_{S}$, the detail loss $\mathcal{L}_{D}$, and the Wasserstein GAN loss, $\mathcal{L}_{adv}$ as:
\vspace{-9pt}
\begin{equation}
\mathcal{L}_{total}=\lambda_{1}\mathcal{L}_{S}+\lambda_{2}\mathcal{L}_{D}+\lambda_{3}\mathcal{L}_{{adv}},
\end{equation}
where $\lambda_1$, $\lambda_2$, and $\lambda_3$ control the effect of the loss functions.

Considering that the proposed dual-stream architecture has two branches, we adopt individual loss functions for each network to update them respectively.
For the multi-color space fusion network, $f_S(\cdot)$, we use the Multi-scale Structural Similarity~\cite{wang2003multiscale} (MS-SSIM) loss, $\mathcal{L}_{\text{MS-SSIM}}$, and $\ell_1$ loss, $\mathcal{L}_{1}$, that can effectively capture the contrast and color errors, respectively:
\vspace{-9pt}

\begin{equation}
\mathcal{L}_{\text{MS-SSIM}}(I_{LL})=1-\text{MS-SSIM}(f_S\left(I_{LL}\right), G_{LL}),
\vspace{-5pt}
\end{equation}
\vspace{-9pt}
\vspace{-9pt}

\begin{equation}
\mathcal{L}_{1}(I_{LL})=\left\|f_S\left(I_{LL}\right)-G_{LL} \right\|_1,
\end{equation}

\noindent
where $G_{LL}$ is the LL image obtained by applying DWT to the ground truth.
The structure loss $\mathcal{L}_{S}$ is then defined by combining $\mathcal{L}_{1}$ and $\mathcal{L}_{\text{MS-SSIM}}$ with the hyperparameter $\alpha$:

\begin{equation}
\mathcal{L}_{S}(I_{LL})=\alpha\cdot \mathcal{L}_{\text{MS-SSIM}}+(1-\alpha )\cdot \mathcal{L}_{1}.
\end{equation}

For the detail enhancement network, $f_D(\cdot)$, we measure the $\ell_2$ error between the estimated detail sub-image, $f_D\left(I_{(i)}\right)$, from $I_{(i)}$ with component $i\in \{LH, HL, HH \}$ and its ground truth, $G_{(i)}$:

\begin{equation}
\mathcal{L}_{D}(I_{(i)})=\left\|f_D\left(I_{(i)}\right)-G_{(i)}\right\|_{2}.
\end{equation}

We further use an adversarial loss, $\mathcal{L}_{adv}$, for the final output, $f(I)$. We employ the loss in Wasserstein GAN~\cite{arjovsky2017wasserstein} that shows stable performance in training. Our network, as a generator, aims to minimize the Wasserstein distance between the real and the generated distribution calculated by the discriminator $C(\cdot)$.

\begin{table}[t]
\vspace{-9pt}
\caption{UIQM, UCIQE, and CIE2000 scores of physics-based (P) and learning-based (L) methods and the average test time per image. The 1st and 2nd best results are in bold and underline, respectively. $\uparrow$: The higher, the better, $\downarrow$: the lower, the better. }
\centering
\renewcommand{\arraystretch}{0.8}
\resizebox{0.95\linewidth}{!}{
\begin{tabular}{l c c c c c c}
\specialrule{1.2pt}{0.2pt}{1pt}
\multicolumn{2}{c}{Methods} & \multicolumn{3}{c}{UIEB~\cite{li2019underwater}} & \multicolumn{2}{c}{ColorChecker~\cite{8058463}}\\
\cmidrule(lr){1-2}
\cmidrule(lr){3-5}
\cmidrule(lr){6-7}
\multicolumn{1}{c}{Name} & P/L  & UIQM $\uparrow$ & UCIQE $\uparrow$ & Time(s) & CIE2000 $\downarrow$ & Time(s)\\
\midrule
\midrule
ULAP~\cite{song2018rapid} & P & 7.213 & 5.832 & 5.09 & 39.98 & 5.69\\
IBLA~\cite{peng2017underwater} & P & 7.385 & 6.129 & 38.71 & 41.35 & 38.54\\
UDCP~\cite{drews2013transmission} & P & 5.648 & \textbf{8.523}  & 15.23 & 39.93 & 17.98  \\
UWCNN~\cite{li2020underwater} & L & 7.791 & 4.769 & 3.31 & \underline{38.21} & 4.40\\
WaterNet~\cite{anwar2018deep} &  L & \textbf{8.818} & 4.465 & 1.56 & 38.63 & 2.25\\
UIE-DAL~\cite{uplavikar2019all} & L & \underline{8.376} & 6.254 & \underline{0.10} & 40.94 & \underline{0.27}\\
Ours & L & 8.032 & \underline{6.341} & \textbf{0.08} & \textbf{37.22} & \textbf{0.26}\\
\specialrule{1.2pt}{0.2pt}{1pt}
\end{tabular}}
\label{tab:quan}
\vspace{-9pt}
\end{table}

\begin{figure}[!t]
\centering
\renewcommand{\arraystretch}{0.5}
\begin{tabular}{c@{\hspace{2pt}}c@{\hspace{2pt}}c@{\hspace{2pt}}c@{\hspace{2pt}}c}
\raisebox{0.3\height}{\rotatebox{90}{\scriptsize Original}}
\begin{overpic}[width=0.22\linewidth,height=0.06\textheight]{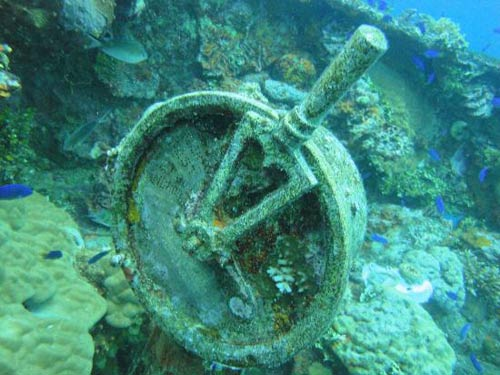}
 \end{overpic}
&\begin{overpic}[width=0.22\linewidth,height=0.06\textheight]{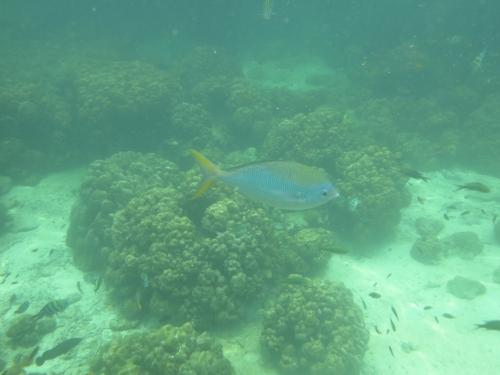}
 \end{overpic}
&\begin{overpic}[width=0.22\linewidth,height=0.06\textheight]{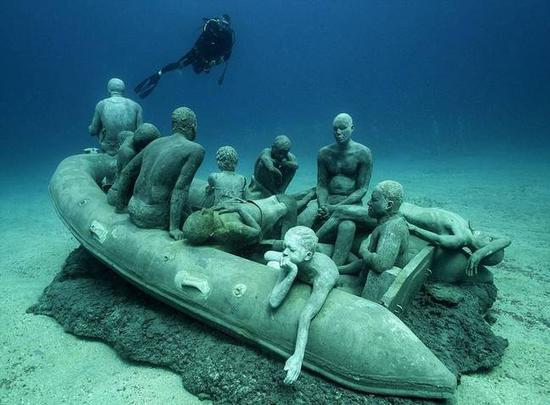}
 \end{overpic}
&\begin{overpic}[width=0.22\linewidth,height=0.06\textheight]{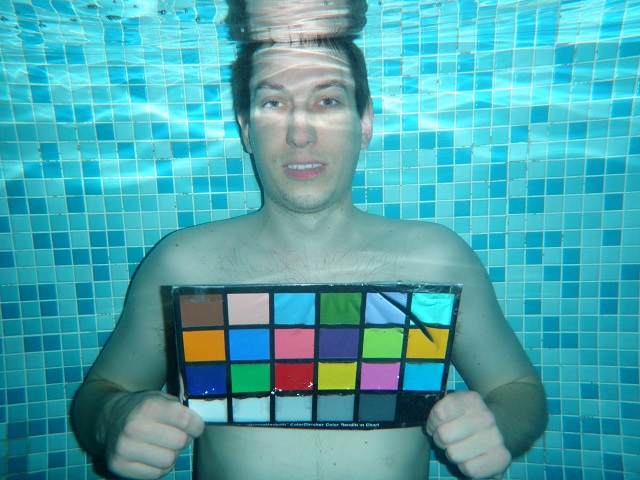}
    \put(65,50){\includegraphics[scale=0.16]{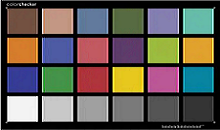}}
\end{overpic}\\

\raisebox{0.5\height}{\rotatebox{90}{\scriptsize ULAP}}
\begin{overpic}[width=0.22\linewidth,height=0.06\textheight]{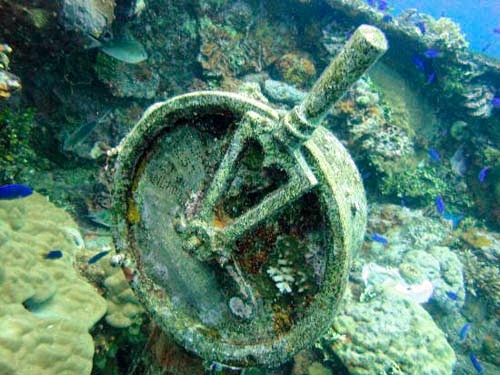}
     \put(1,62){\scriptsize \contour{black}{\protect\textcolor{white}{3.42/4.78}}}
 \end{overpic}
&\begin{overpic}[width=0.22\linewidth,height=0.06\textheight]{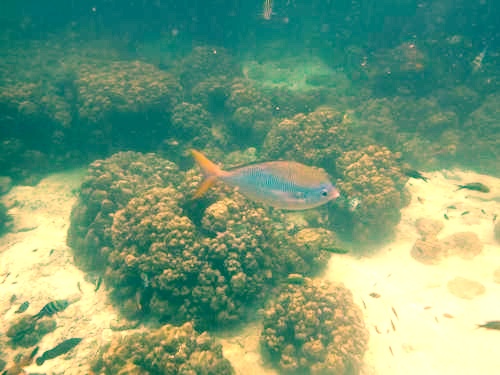}
     \put(1,62){\scriptsize \contour{black}{\protect\textcolor{white}{7.82/2.99}}}
 \end{overpic}
&\begin{overpic}[width=0.22\linewidth,height=0.06\textheight]{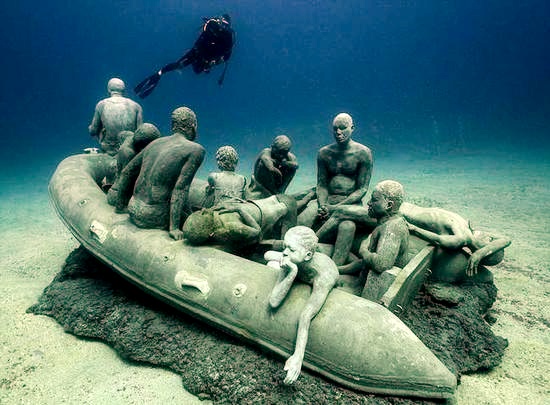}
     \put(1,62){\scriptsize \contour{black}{\protect\textcolor{white}{4.24/3.63}}}
 \end{overpic}
& \begin{overpic}[width=0.22\linewidth,height=0.06\textheight]{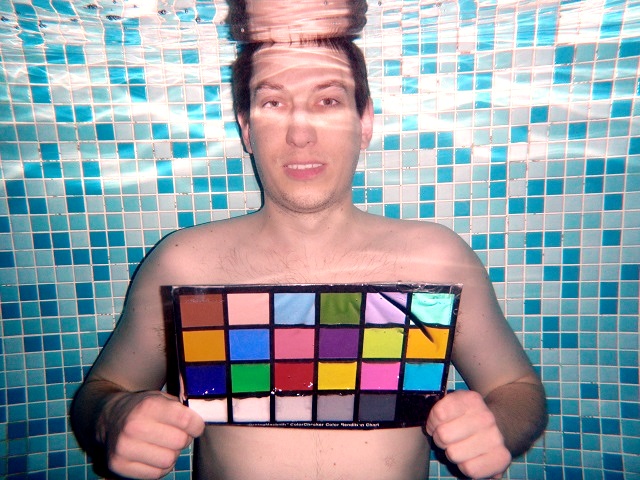}
     \put(70,2){\scriptsize \contour{black}{\protect\textcolor{white}{40.36}}}
     \put(1,62){\scriptsize \contour{black}{\protect\textcolor{black}{8.21/6.84}}}
 \end{overpic}\\

\raisebox{0.7\height}{\rotatebox{90}{\scriptsize IBLA}}
\begin{overpic}[width=0.22\linewidth,height=0.06\textheight]{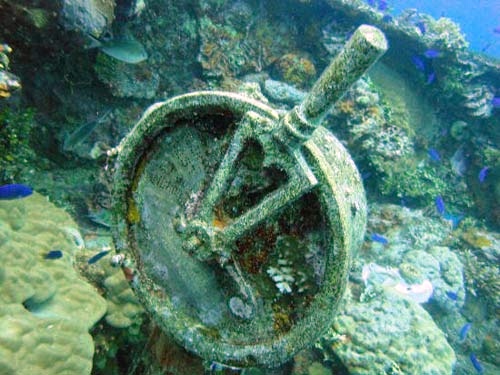}
     \put(1,62){\scriptsize \contour{black}{\protect\textcolor{white}{3.83/4.44}}}
 \end{overpic}
&\begin{overpic}[width=0.22\linewidth,height=0.06\textheight]{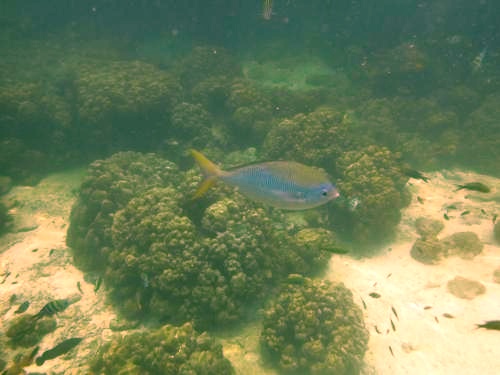}
     \put(1,62){\scriptsize \contour{black}{\protect\textcolor{white}{8.79/3.63}}}
 \end{overpic}
&\begin{overpic}[width=0.22\linewidth,height=0.06\textheight]{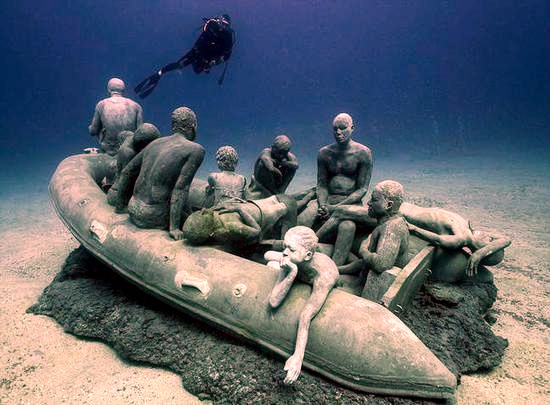}
     \put(1,62){\scriptsize \contour{black}{\protect\textcolor{white}{9.61/5.41}}}
 \end{overpic}
& \begin{overpic}[width=0.22\linewidth,height=0.06\textheight]{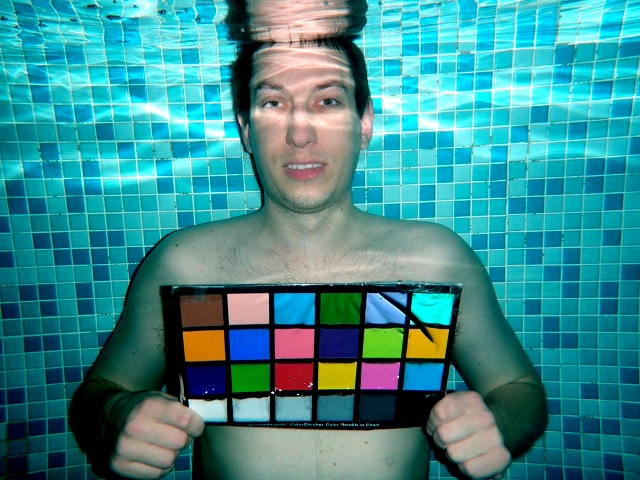}
     \put(70,2){\scriptsize \contour{black}{\protect\textcolor{white}{41.41}}}
     \put(1,62){\scriptsize \contour{black}{\protect\textcolor{white}{6.07/6.98}}}
 \end{overpic}\\

\raisebox{0.5\height}{\rotatebox{90}{\scriptsize UDCP}}
\begin{overpic}[width=0.22\linewidth,height=0.06\textheight]{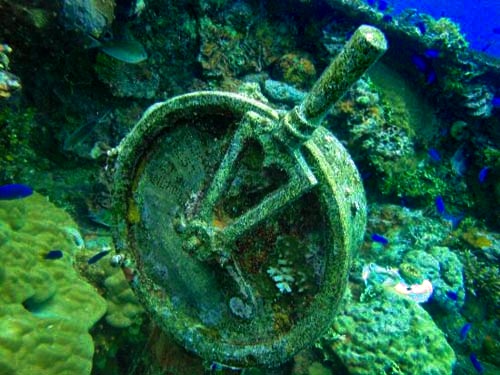}
     \put(1,62){\scriptsize \contour{black}{\protect\textcolor{white}{2.02/9.68}}}
 \end{overpic}
&\begin{overpic}[width=0.22\linewidth,height=0.06\textheight]{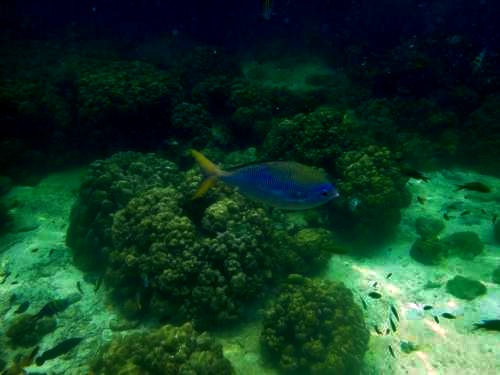}
     \put(1,62){\scriptsize \contour{black}{\protect\textcolor{white}{2.06/8.42}}}
 \end{overpic}
&\begin{overpic}[width=0.22\linewidth,height=0.06\textheight]{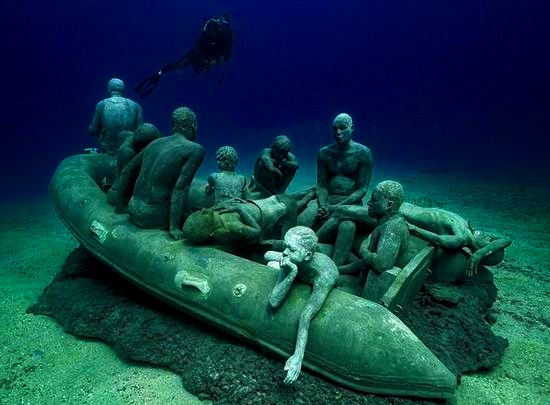}
     \put(1,62){\scriptsize \contour{black}{\protect\textcolor{white}{2.61/7.36}}}
 \end{overpic}
& \begin{overpic}[width=0.22\linewidth,height=0.06\textheight]{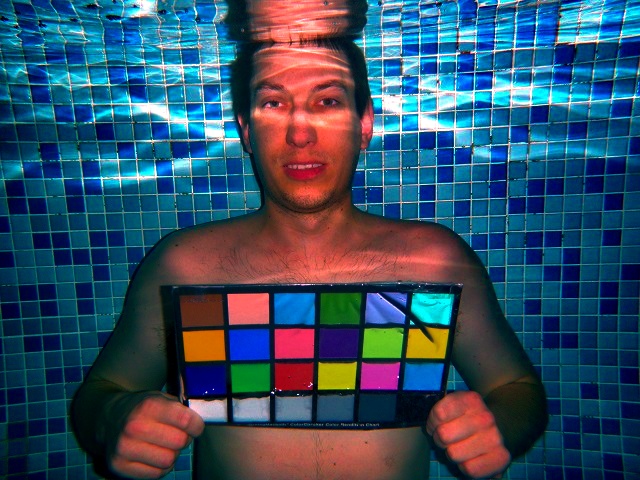}
     \put(70,2){\scriptsize \contour{black}{\protect\textcolor{white}{39.93}}}
     \put(1,62){\scriptsize \contour{black}{\protect\textcolor{white}{5.80/9.31}}}
 \end{overpic}\\

\raisebox{0.2\height}{\rotatebox{90}{\scriptsize UWCNN}}
\begin{overpic}[width=0.22\linewidth,height=0.06\textheight]{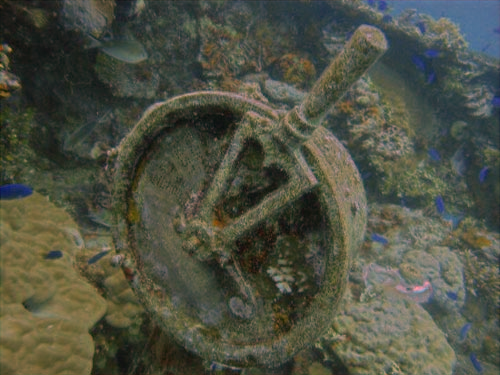}
     \put(1,62){\scriptsize \contour{black}{\protect\textcolor{white}{10.02/3.43}}}
 \end{overpic}
&\begin{overpic}[width=0.22\linewidth,height=0.06\textheight]{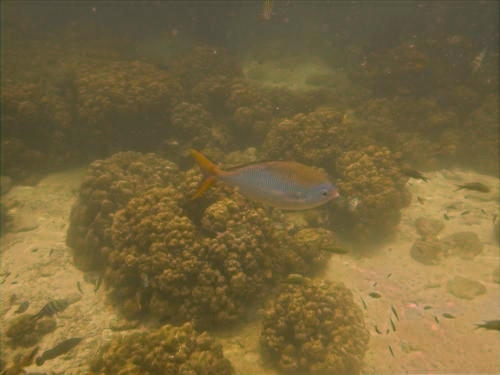}
     \put(1,62){\scriptsize \contour{black}{\protect\textcolor{white}{7.83/4.44}}}
 \end{overpic}
&\begin{overpic}[width=0.22\linewidth,height=0.06\textheight]{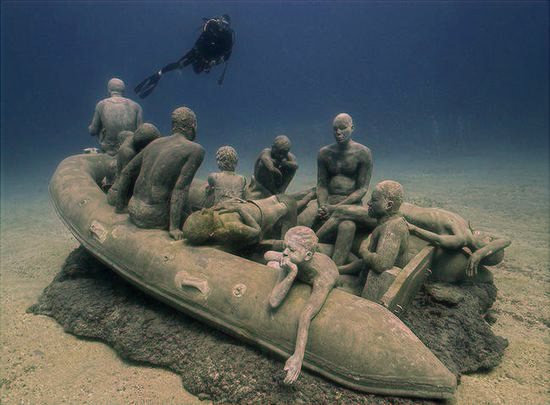}
     \put(1,62){\scriptsize \contour{black}{\protect\textcolor{white}{11.23/2.81}}}
 \end{overpic}
& \begin{overpic}[width=0.22\linewidth,height=0.06\textheight]{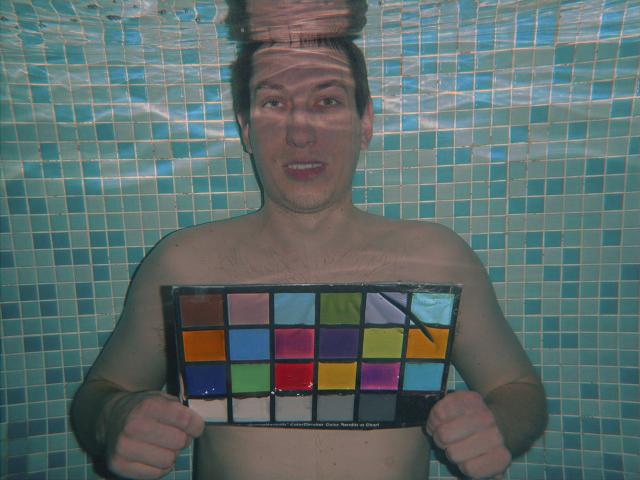}
     \put(70,2){\scriptsize\contour{black}{\protect\textcolor{white}{39.42}}}
     \put(1,62){\scriptsize \contour{black}{\protect\textcolor{white}{9.95/3.84}}}
 \end{overpic}\\

\raisebox{0.2\height}{\rotatebox{90}{\scriptsize WaterNet}}
\begin{overpic}[width=0.22\linewidth,height=0.06\textheight]{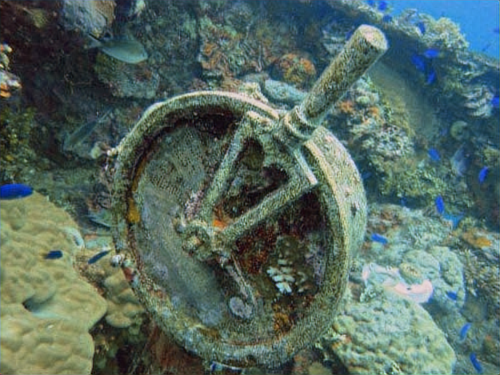}
     \put(1,62){\scriptsize \contour{black}{\protect\textcolor{white}{4.59/4.34}}}
 \end{overpic}
&\begin{overpic}[width=0.22\linewidth,height=0.06\textheight]{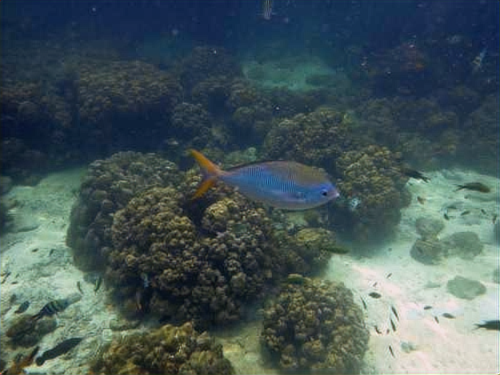}
     \put(1,62){\scriptsize \contour{black}{\protect\textcolor{white}{3.72/3.49}}}
 \end{overpic}
&\begin{overpic}[width=0.22\linewidth,height=0.06\textheight]{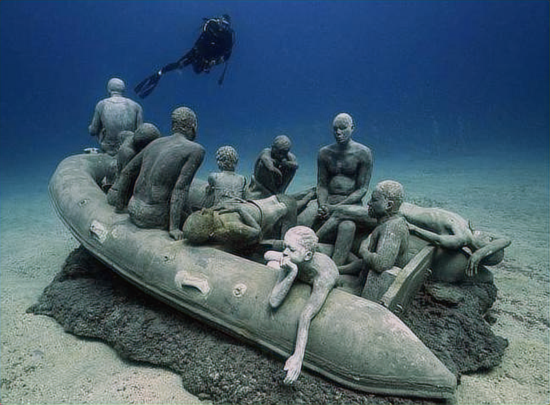}
     \put(1,62){\scriptsize \contour{black}{\protect\textcolor{white}{5.78/6.09}}}
 \end{overpic}
& \begin{overpic}[width=0.22\linewidth,height=0.06\textheight]{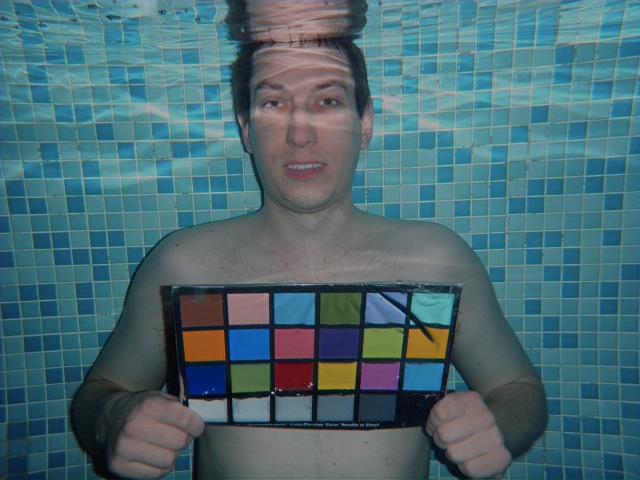}
     \put(70,2){\scriptsize \contour{black}{\protect\textcolor{white}{38.07}}}
     \put(1,62){\scriptsize \contour{black}{\protect\textcolor{white}{6.09/4.50}}}
 \end{overpic}\\

\raisebox{0.2\height}{\rotatebox{90}{\scriptsize UIE-DAL}}
\begin{overpic}[width=0.22\linewidth,height=0.06\textheight]{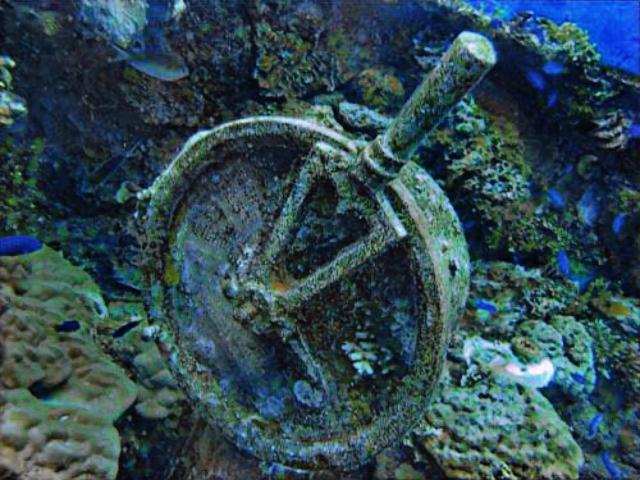}
     \put(1,62){\scriptsize \contour{black}{\protect\textcolor{white}{3.00/3.88}}}
 \end{overpic}
&\begin{overpic}[width=0.22\linewidth,height=0.06\textheight]{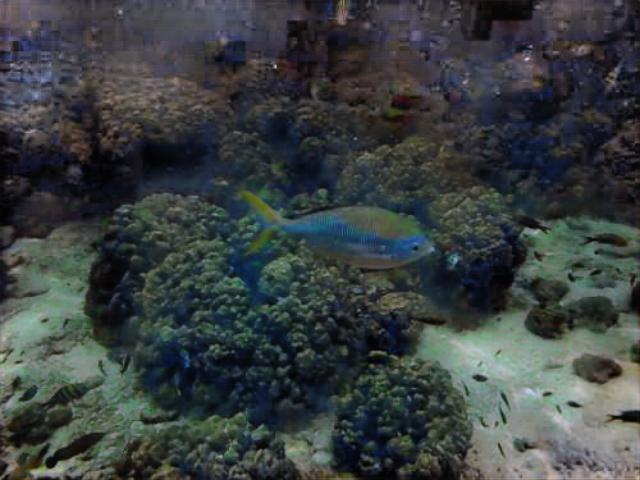}
     \put(1,62){\scriptsize \contour{black}{\protect\textcolor{white}{10.57/3.21}}}
 \end{overpic}
&\begin{overpic}[width=0.22\linewidth,height=0.06\textheight]{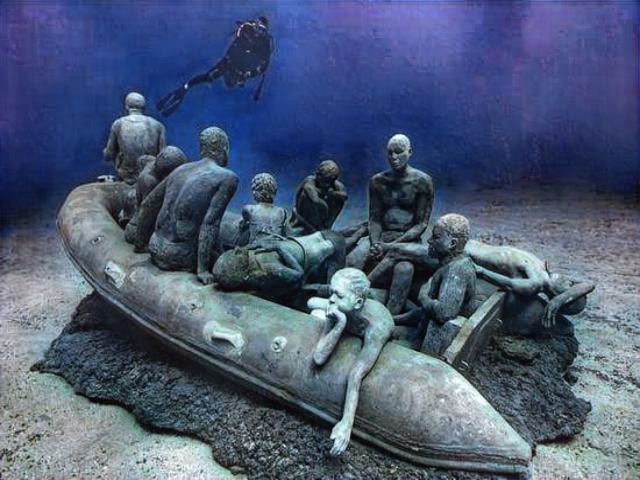}
     \put(1,62){\scriptsize \contour{black}{\protect\textcolor{white}{4.03/8.68}}}
 \end{overpic}
& \begin{overpic}[width=0.22\linewidth,height=0.06\textheight]{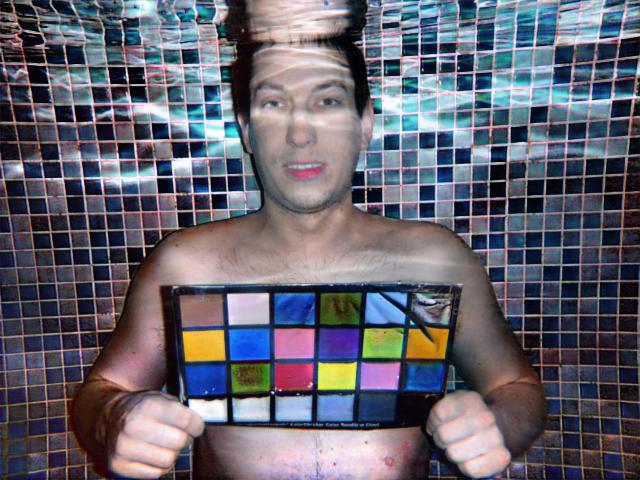}
     \put(70,2){\scriptsize \contour{black}{\protect\textcolor{white}{40.25}}}
     \put(1,62){\scriptsize \contour{black}{\protect\textcolor{white}{10.69/6.41}}}
 \end{overpic}\\

\raisebox{0.9\height}{\rotatebox{90}{\scriptsize Ours}}
\begin{overpic}[width=0.22\linewidth,height=0.06\textheight]{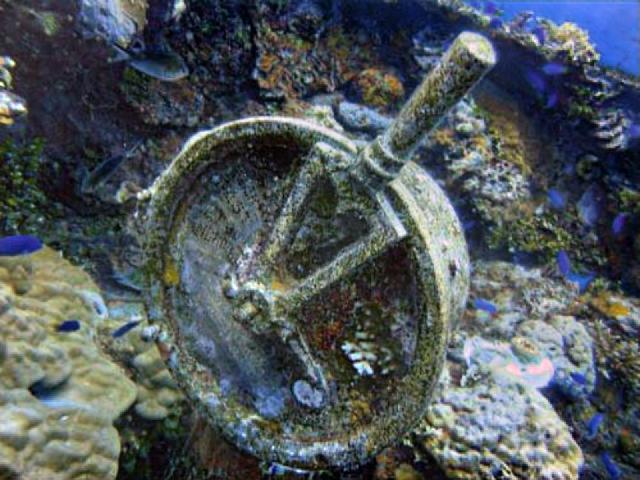}
     \put(1,62){\scriptsize \contour{black}{\protect\textcolor{white}{3.23/5.48}}}
 \end{overpic}
&\begin{overpic}[width=0.22\linewidth,height=0.06\textheight]{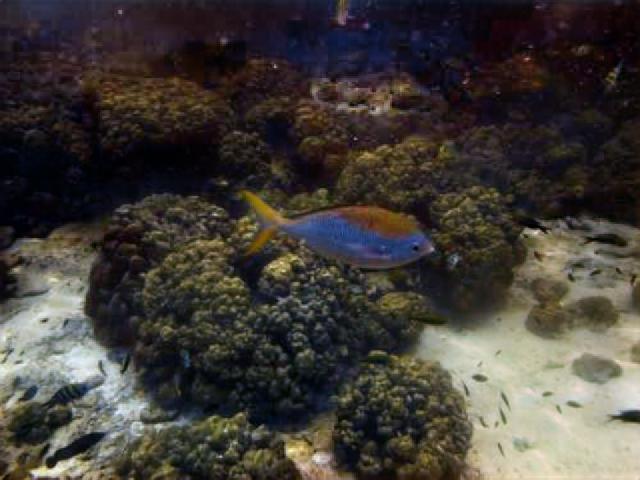}
     \put(1,62){\scriptsize \contour{black}{\protect\textcolor{white}{11.42/3.45}}}
 \end{overpic}
&\begin{overpic}[width=0.22\linewidth,height=0.06\textheight]{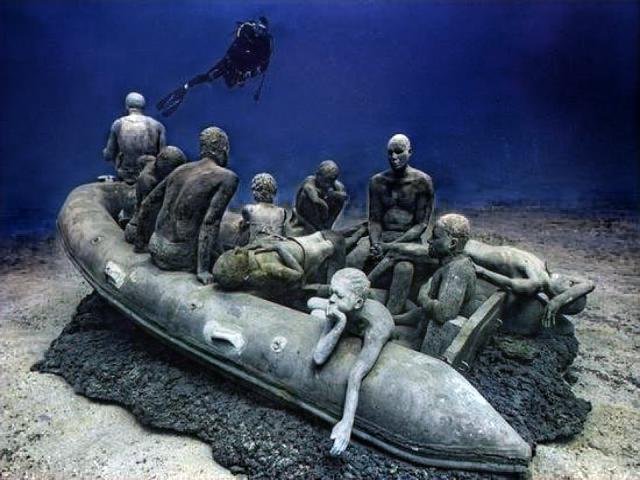}
     \put(1,62){\scriptsize \contour{black}{\protect\textcolor{white}{4.23/8.77}}}
 \end{overpic}
& \begin{overpic}[width=0.22\linewidth,height=0.06\textheight]{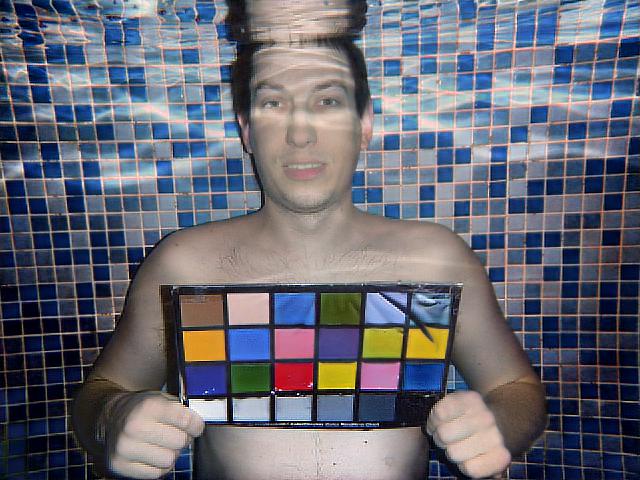}
    \put(70,2){\scriptsize \contour{black}{\protect\textcolor{white}{36.98}}}
     \put(1,62){\scriptsize \contour{black}{\protect\textcolor{white}{10.68/6.70}}}
 \end{overpic}\\

\footnotesize{(a)}
& \footnotesize{(b)}
& \footnotesize{(c)}
& \footnotesize{(d)}\\
\end{tabular}
\vspace{-9pt}
\caption{Visual samples of underwater image enhancement from (a-c) UIEB~\cite{li2019underwater} and (d) ColorChecker~\cite{8058463}. UIQM and UCIQE scores (the higher, the better) are presented on the top-left and the CIE2000 score (the lower, the better) is presented on the bottom-right in (d). }
\label{fig:qualtive}
\vspace{-9pt}
\end{figure}
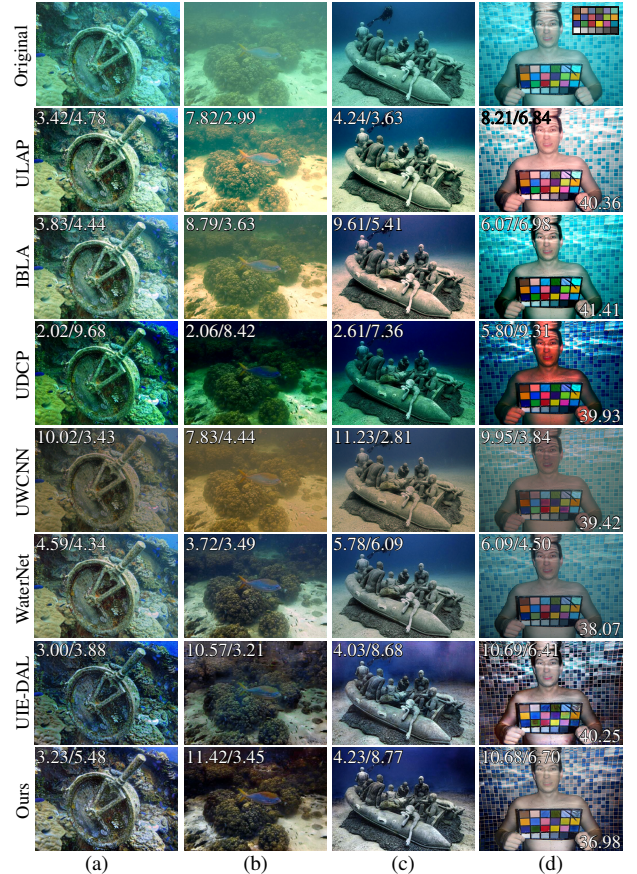

\section{Validation}

\subsection{Setup}

Our network, implemented with PyTorch, was trained and tested on Intel Core i5-7200 CPU and NVIDIA Geforce RTX 2070 GPU.
We used RMSProp~\cite{tieleman2012lecture} as an optimizer with a batch size of 4 and set learning rate to 0.0005 for $f_S(\cdot)$, and 0.00002 for $f_D(\cdot)$. The weights, $\lambda_{1}$, $\lambda_{2}$, $\lambda_{3}$, $\alpha$, were set to 0.5, 1, 1, 0.5, respectively.  We first trained the dual-stream network and then added the discriminator into the training process. The discriminator was trained 5 epochs for each epoch of the training for the dual-stream network.

Due to the lack of access to the ground truth for real underwater images, we followed the training protocol in~\cite{anwar2018deep} to train our model. We used a synthetic underwater dataset produced from NYU-v2 dataset~\cite{silberman2012indoor} with 1449 indoor images. We augmented each image to 36 images with 6 different Jerlov water types and 6 different levels of background light, and selected 20k images as our training dataset. For testing, we used both synthetic and real-world underwater images, including the NYU-v2 with 3000 synthetic images, the UIEB dataset~\cite{li2019underwater} with 890 real underwater images, and the ColorChecker dataset~\cite{8058463} collected by Olympus Tough 6000 with 7 images.

For quantitative comparisons of underwater images, we used UIQM~\cite{panetta2015human}, UCIQE~\cite{yang2015underwater}, and CIE2000~\cite{sharma2005ciede2000} metrics. UIQM considers the colorfulness, sharpness, and contrast while UCIQE evaluates the chroma, saturation, and contrast of images. CIE2000 measures the color difference between two images, which can be used for the color checker images. We also employ the SSIM to evaluate the effectiveness of our network in the ablation studies, which can be measured on the synthetic dataset~\cite{silberman2012indoor} with the ground-truth images.

\subsection{Results and comparisons}
\label{ssc:Results and Comparisons}
Fig.~\ref{fig:qualtive} shows visual results generated by existing methods and ours on the UIEB dataset~\cite{anwar2018deep} and the underwater ColorChecker~\cite{8058463}. Physics-based methods, ULAP~\cite{song2018rapid}, IBLA~\cite{peng2017underwater}, UDCP~\cite{drews2013transmission} improve the brightness and contrast well, but fail to remove severe color cast and blurring. Existing learning-based methods, Water-Net~\cite{anwar2018deep}, UWCNN~\cite{li2020underwater}, UIE-DAL~\cite{uplavikar2019all} can deal with the color cast and blurring issues, but the artifacts are observed in the color checkers (UWCNN, UIE-DAL) or the contrast is low (UWCNN, WaterNet). Our network produces the images by effectively removing the color cast with the improved details while avoiding unnatural artifacts.

\begin{table}[t]
\caption{Ablation study on the loss functions. }
\centering
\renewcommand{\arraystretch}{0.8}
\resizebox{0.85\linewidth}{!}{
\begin{tabular}{c c c c c c}
\specialrule{1.2pt}{0.2pt}{1pt}
\multicolumn{3}{c}{Loss functions} & \multicolumn{1}{c}{NYU-v2~\cite{silberman2012indoor}} & \multicolumn{2}{c}{UIEB~\cite{li2019underwater}} \\
\cmidrule(lr){1-3}
\cmidrule(lr){4-4}
\cmidrule(lr){5-6}
$\mathcal{L}_{\text{MS-SSIM}}$ & $\mathcal{L}_{1}$ & $\mathcal{L}_{Adv}$ & SSIM $\uparrow$ & UIQM $\uparrow$ & UCIQE $\uparrow$ \\
\midrule
\midrule
\checkmark & & & 0.882 & 6.551 & 6.084 \\
& \checkmark & & 0.884 & 6.604 & \textbf{6.812}\\
\checkmark & \checkmark & & 0.896 & 7.341 & 6.061\\
\checkmark & \checkmark & \checkmark & \textbf{0.918} & \textbf{8.032} & 6.341\\
\specialrule{1.2pt}{0.2pt}{1pt}
\end{tabular}}
\label{tab:loss}
\vspace{-9pt}
\end{table}
\begin{figure}[t]
\centering
\begin{tabular}{c@{\hspace{2pt}}c@{\hspace{2pt}}c@{\hspace{2pt}}c}
\includegraphics[width=0.22\linewidth,height=0.06\textheight]{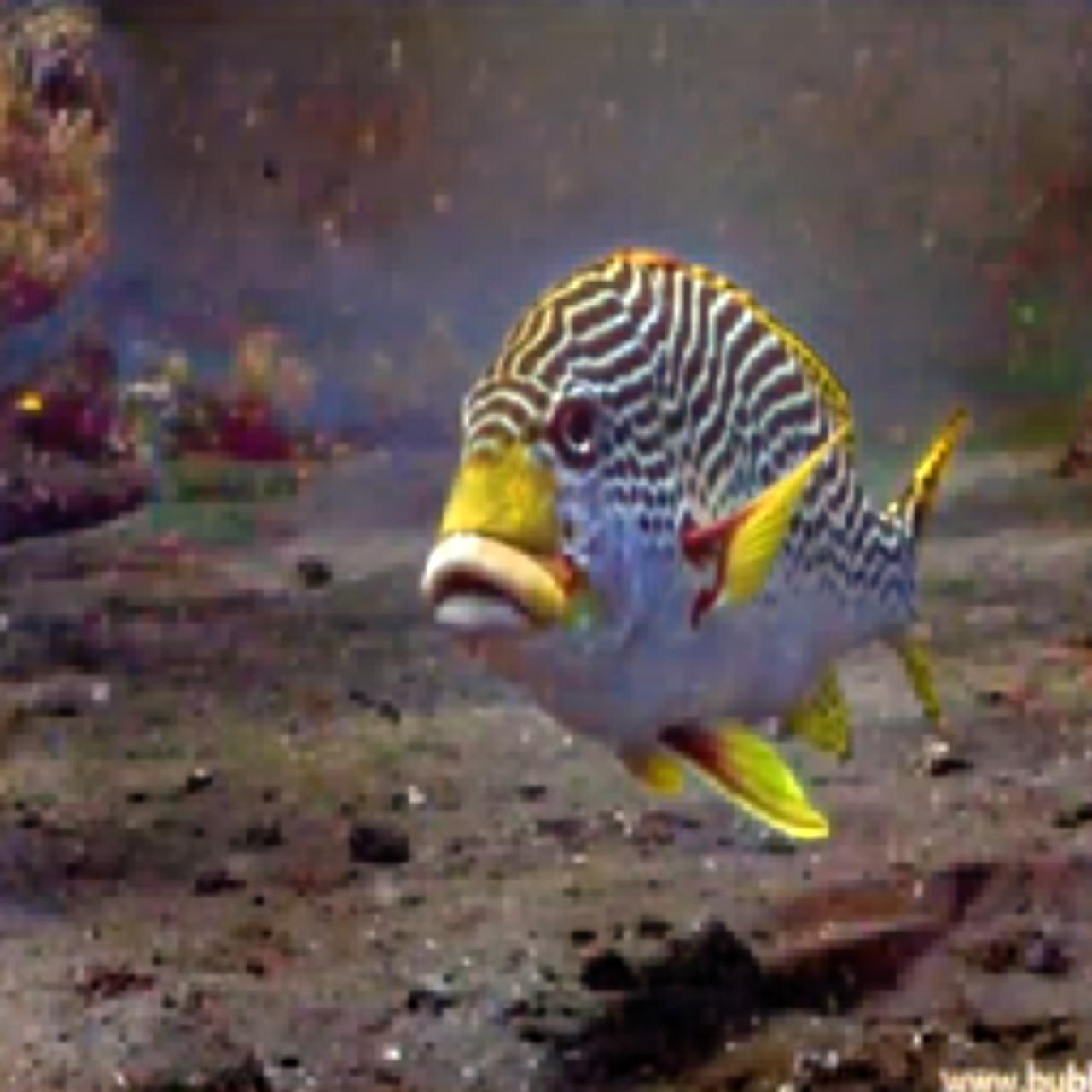}
&\includegraphics[width=0.22\linewidth,height=0.06\textheight]{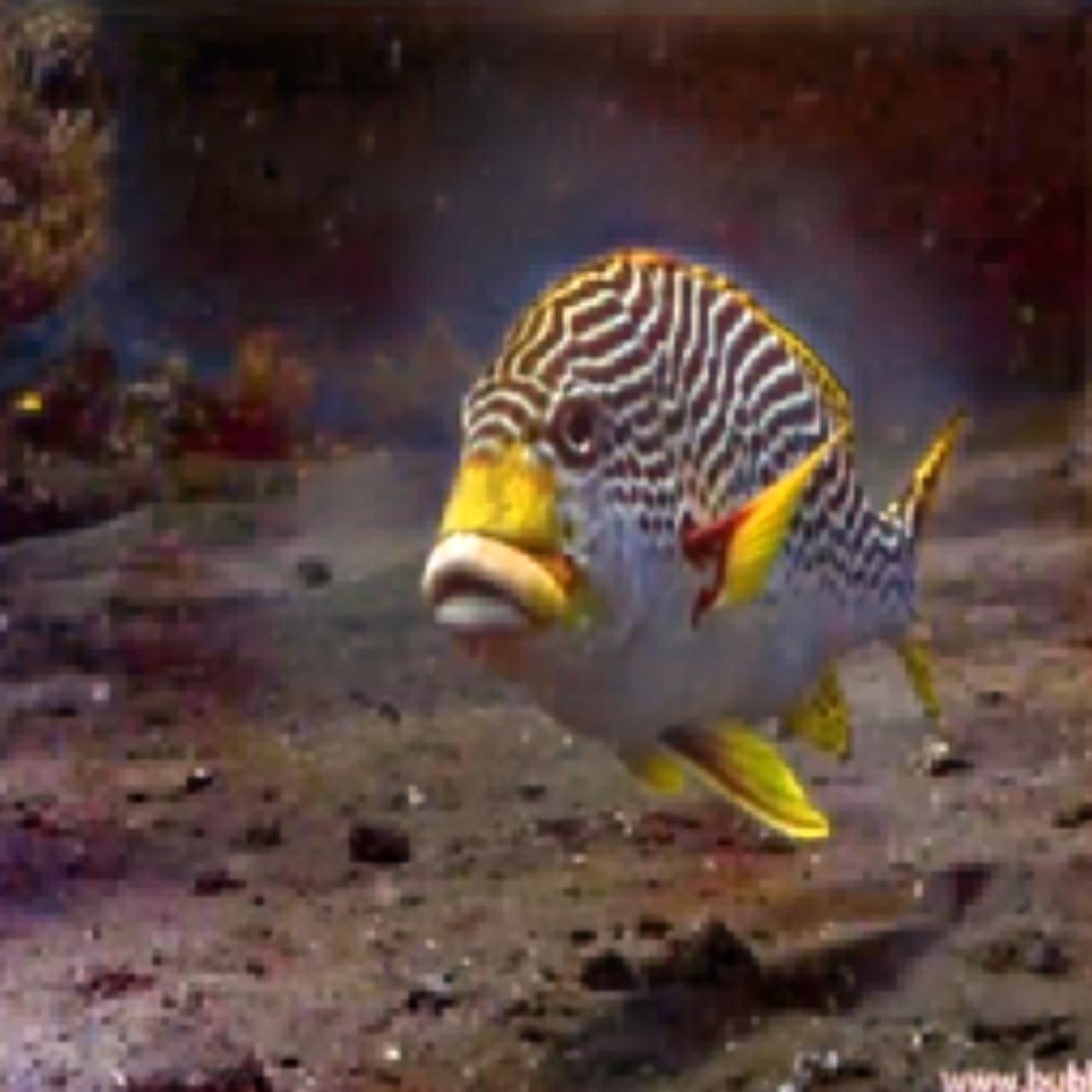}
&\includegraphics[width=0.22\linewidth,height=0.06\textheight]{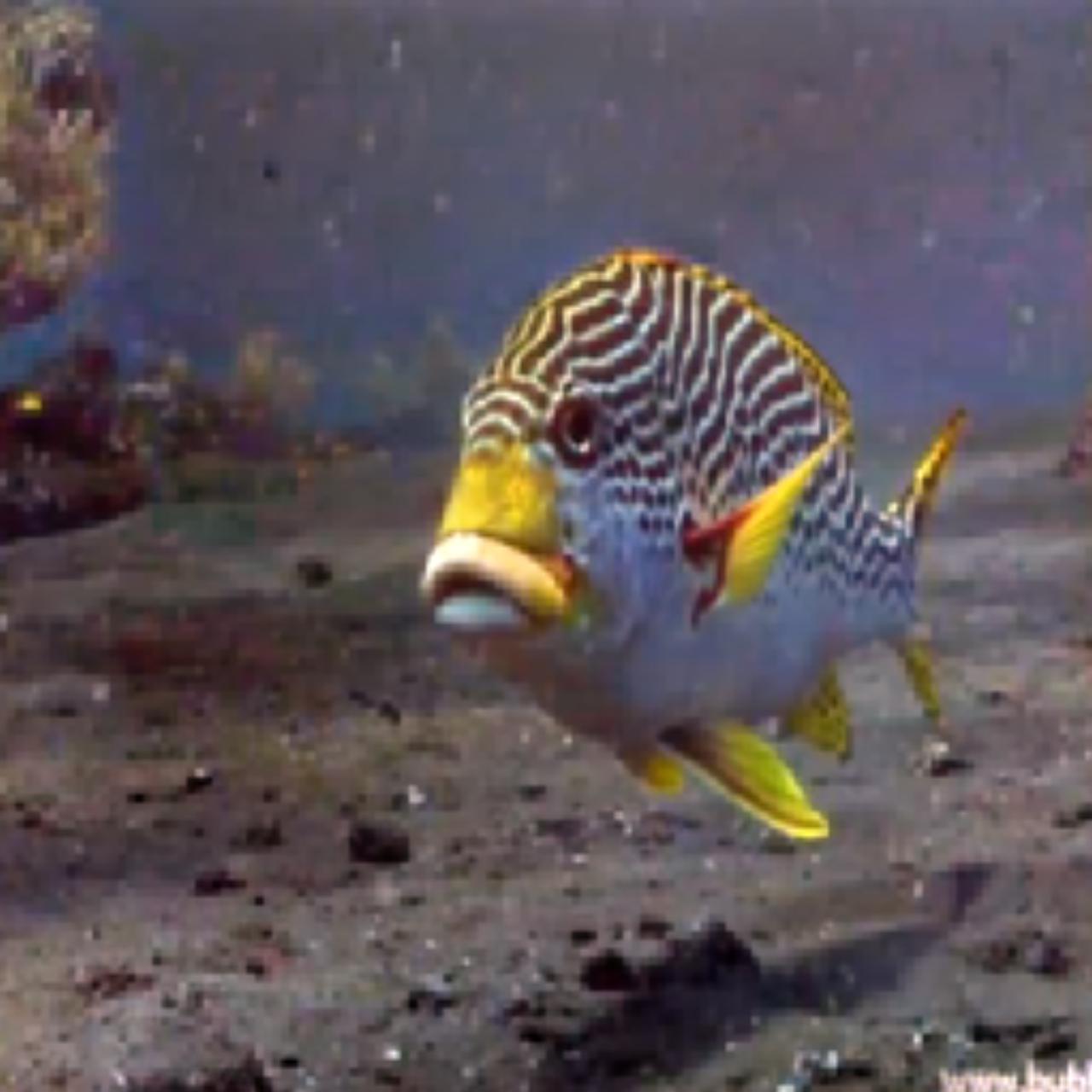}
&\includegraphics[width=0.22\linewidth,height=0.06\textheight]{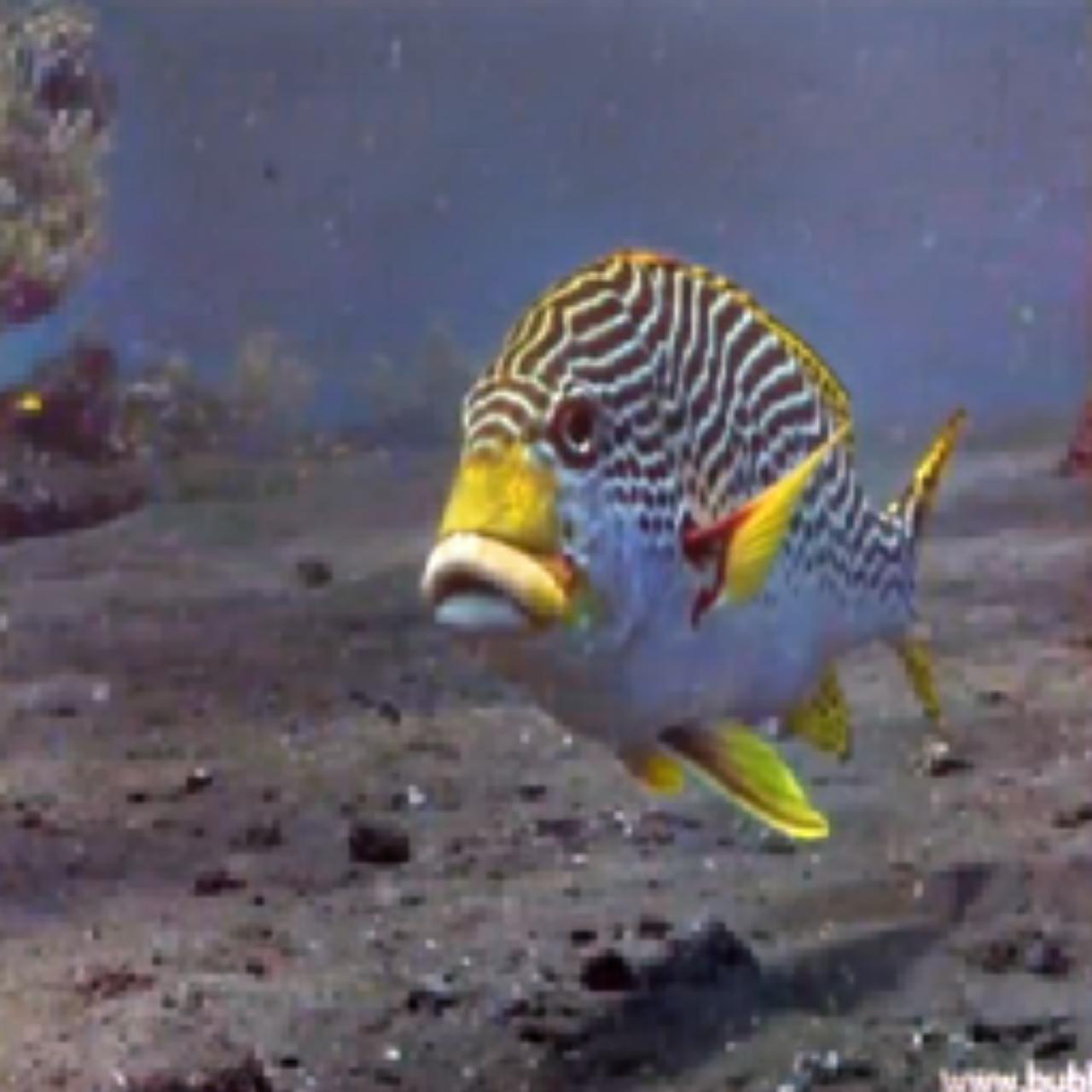}\\
\footnotesize{$\mathcal{L}_1$} & \footnotesize{$\mathcal{L}_{\text{MS-SSIM}}$} & \footnotesize{$\mathcal{L}_1$+$\mathcal{L}_{\text{MS-SSIM}}$} & \footnotesize{$\mathcal{L}_{total}$}\\
\end{tabular}
\vspace{-9pt}
\caption{Results from the models trained with different loss functions. }
\label{fig:abl_loss}
\vspace{-9pt}
\end{figure}

Table~\ref{tab:quan} shows the quantitative results, where the learning-based approaches generally show better scores than the physics-based approaches. For UCIQE scores, except for UDCP, ours shows higher scores than other methods. Although UDCP has a higher UCIQE value, the visual results show low intensities in green and blue channels (Fig.~\ref{fig:qualtive}(a)-(c)), which makes the global tone relatively red (Fig.~\ref{fig:qualtive}(d)). The physics-based methods show limited performance in heavily blurred images, e.g., Fig.~\ref{fig:qualtive}(b), thus resulting in a low sharpness. Our method outperforms others in detail enhancement, which makes our UIQM values generally higher in heavily blurred images.
To better compare the color correction, we analyze the performance of color constancy using the images taken underwater with the standard color checker~\cite{8058463}. Our model obtains the lowest CIE2000 score, indicating that the colors in the color checker are the closest color to the real ones.
We also measure the average testing time for each method on the UIEB dataset and ColorChecker. The physics-based methods take a longer time to enhance the image, while the learning-based methods take less time. Our network, thanks to image decomposition, uses the smaller size of inputs and thus shows low computational complexity.

\subsection{Ablation study}
\textbf{Loss functions.}
We examine the loss functions by training $f_S(\cdot)$ with different settings: $\mathcal{L}_1$, $\mathcal{L}_{\text{MS-SSIM}}$, and $\mathcal{L}_1$+$\mathcal{L}_{\text{MS-SSIM}}$. Table~\ref{tab:loss} and Fig.~\ref{fig:abl_loss} show the results of the ablation study.
The results from $\mathcal{L}_1$+$\mathcal{L}_{\text{MS-SSIM}}$ shows less visual artifacts than $\mathcal{L}_1$ and $\mathcal{L}_{\text{MS-SSIM}}$. The results from $\mathcal{L}_{\text{MS-SSIM}}$ show high contrast, which leads to a high UCIQE score. However, the contrast enhancement is only focused near the object. Finally, the result from $\mathcal{L}_{\text{total}}$ shows a less noisy and more colorful output. The quantitative results also show that $\mathcal{L}_{\text{total}}$ achieves the highest SSIM and UIQM.

\begin{table}[t!]
\caption{Ablation study on the effect of discrete wavelet transform (DWT), $f_D(\cdot)$, multi-color space (MCS), and generative adversarial network (GAN).}
\centering
\renewcommand{\arraystretch}{0.8}
\resizebox{0.95\linewidth}{!}{
\begin{tabular}{c c c cc c c}
\specialrule{1.2pt}{0.2pt}{1pt}
\multicolumn{4}{c}{Modules} & \multicolumn{1}{c}{NYU-v2~\cite{silberman2012indoor}} & \multicolumn{2}{c}{UIEB~\cite{li2019underwater}} \\
\cmidrule(lr){1-4}
\cmidrule(lr){5-5}
\cmidrule(lr){6-7}
{DWT} &{$f_D(\cdot)$} &{MCS} &{GAN} & SSIM $\uparrow$ & UIQM $\uparrow$ & UCIQE $\uparrow$\\
\midrule
\midrule
\checkmark &  & & &  0.841 & 7.006 & 5.283 \\
&  \checkmark & & &  0.852 &  7.125 & 5.647\\
\checkmark &  \checkmark & & &  0.854 & 7.095 & 5.426 \\
\checkmark &  \checkmark & \checkmark & &  0.872 & 7.341 & 6.061 \\
\checkmark & \checkmark & \checkmark & \checkmark &  \textbf{0.911} & \textbf{8.032} & \textbf{6.341}\\
\specialrule{1.2pt}{0.2pt}{1pt}
\end{tabular}}
\label{tab:abl mod}
\end{table}

\begin{figure}[!t]
\centering
\renewcommand{\arraystretch}{0.5}
\begin{tabular}{c@{\hspace{2pt}}c@{\hspace{2pt}}c}
\includegraphics[width=0.22\linewidth,height=0.06\textheight]{figures/results/1_original.png}
&\includegraphics[width=0.22\linewidth,height=0.06\textheight]{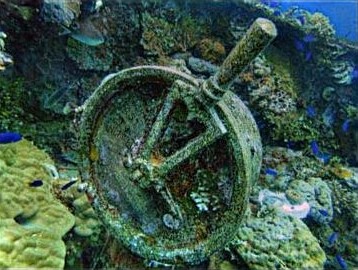}
&\includegraphics[width=0.22\linewidth,height=0.06\textheight]{figures/results/1_ours.png}\\

\includegraphics[width=0.22\linewidth,height=0.06\textheight]{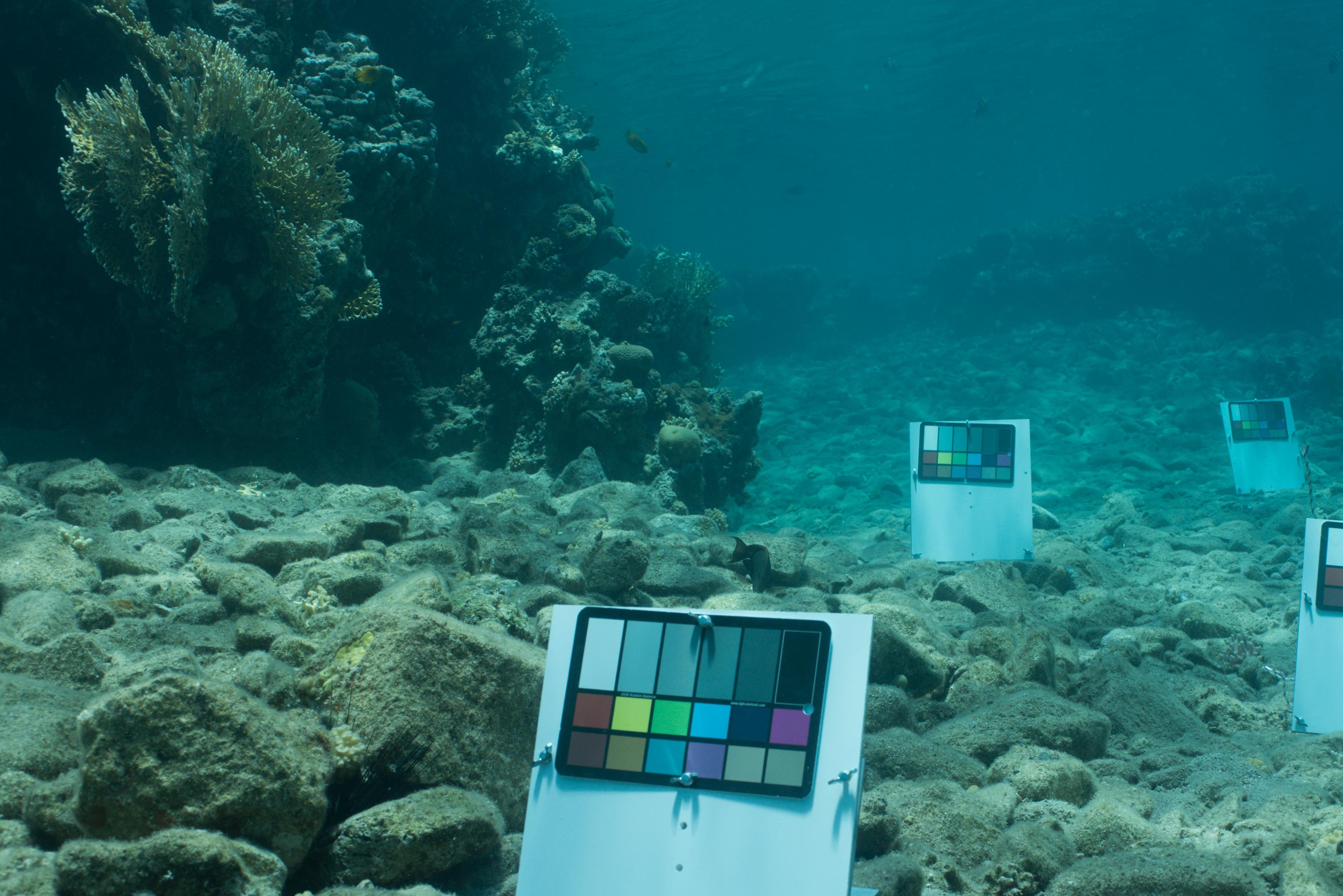}
&\includegraphics[width=0.22\linewidth,height=0.06\textheight]{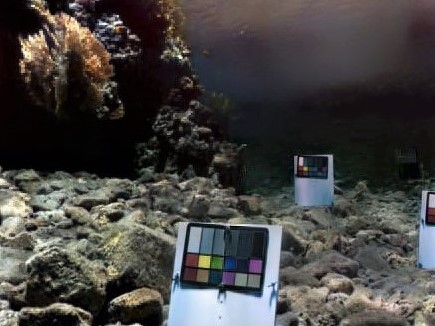}
&\includegraphics[width=0.22\linewidth,height=0.06\textheight]{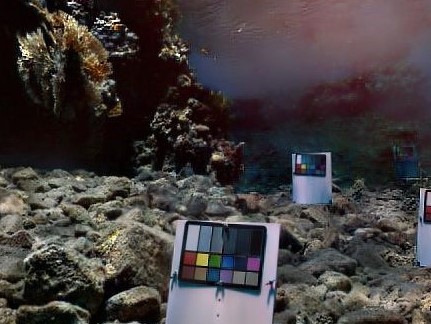}\\

\includegraphics[width=0.22\linewidth,height=0.06\textheight]{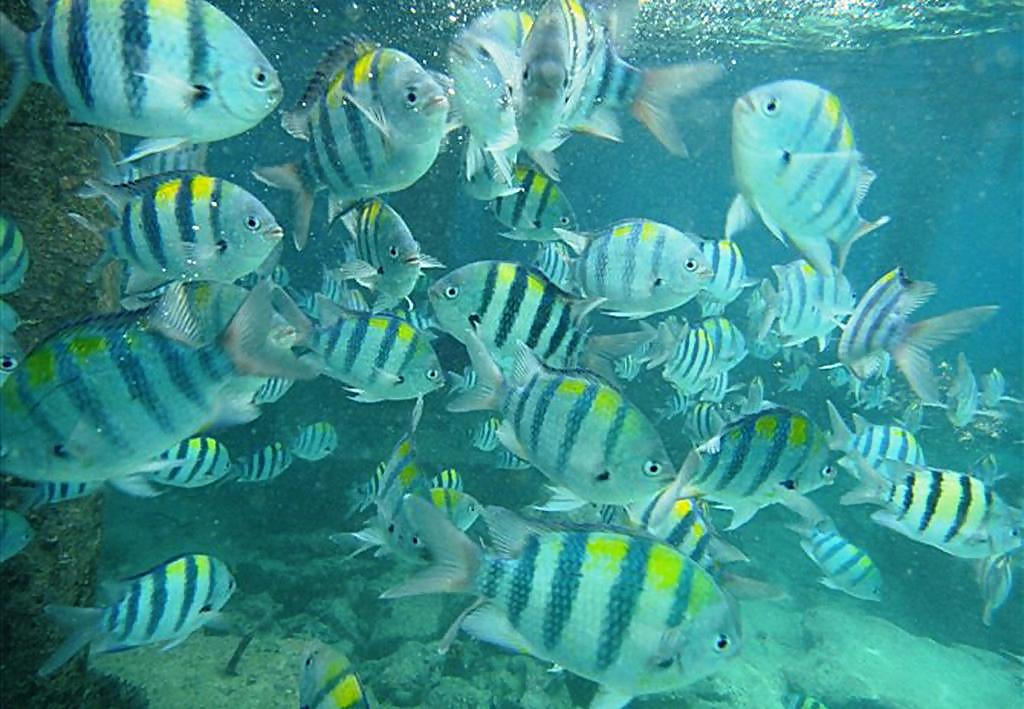}
&\includegraphics[width=0.22\linewidth,height=0.06\textheight]{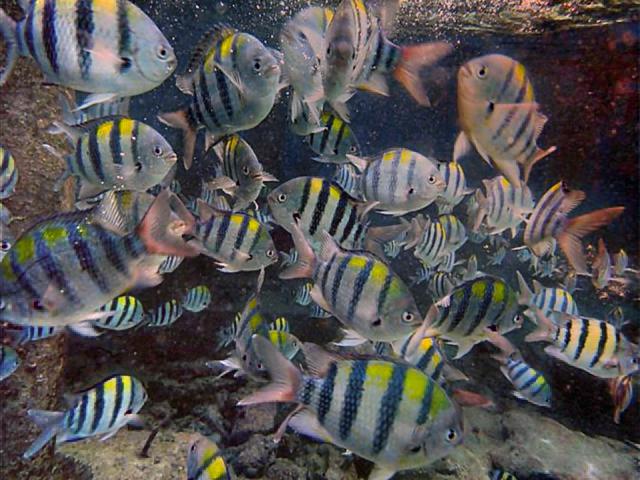}
&\includegraphics[width=0.22\linewidth,height=0.06\textheight]{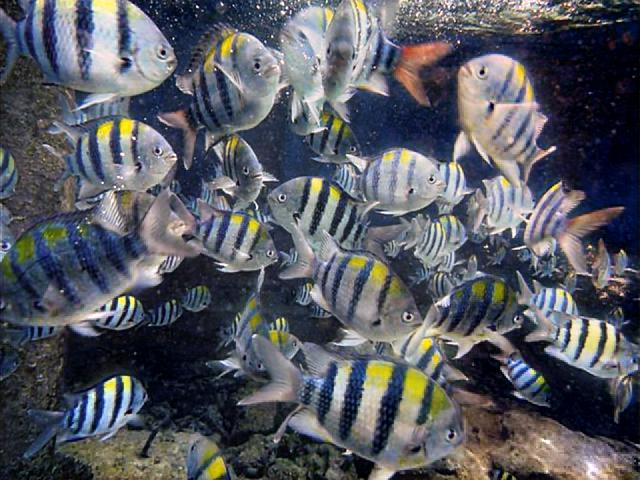}\\
\footnotesize{(a)} & \footnotesize{(b)} & \footnotesize{(c)}\\
\end{tabular}
\vspace{-9pt}
\caption{Ablation study for each module. (a) Inputs, (b,c) the results without and with (top)  discrete wavelet transform, (middle) $f_D(\cdot)$, and (bottom) multi-color space. }
\label{fig:abl_mod}
\vspace{-9pt}
\end{figure}

\noindent
\textbf{Network configurations.}
We examine each component of our model. Fig.~\ref{fig:abl_mod} and Table~\ref{tab:abl mod} show the results of the ablation study.
We compare our results with a dual-stream network~\cite{pan2018learning} that uses an RGB image for both sub-networks. As shown in the first row of Fig.~\ref{fig:abl_mod}, our network can address the underwater image enhancement better by separately addressing the color correction and detail enhancement.
To see the effect of $f_D(\cdot)$ for detail enhancement, we compare our results, $f(I)$, with the images reconstructed by $f_S(I_{LL})$ and the original details, $I_{LH}$, $I_{HL}$, and $I_{HH}$. We can observe the advantage of $f_D(\cdot)$ as shown in the 2nd row of Fig.~\ref{fig:abl_mod} and Table~\ref{tab:abl mod}. We further test $f_S(\cdot)$ by replacing this model with  the vanilla U-net~\cite{ronneberger2015u}, which shows the advantage of multi-color space in $f_S(\cdot)$ with better contrast and color appearance.

\section{Conclusion}
\label{sec:concl}
We presented a method to decouple the various artifacts in underwater image enhancement by using DWT, which can separately perform color correction and detail enhancement with the proposed dual-stream network. The key idea is that the sub-band image with low-frequency contains the color cast and the rest of sub-band images with high-frequency contain blurry details. We showed that our model can effectively remove the color cast and improve the blurry details. The test time is reduced by decomposing the input into smaller sizes. Future work includes a subjective evaluation of our results and investigating a metric for visual quality assessment of underwater images.

\end{document}